\pdfoutput=1
\documentclass[11pt]{article}
\usepackage{emnlp2021}
\usepackage{times}
\usepackage{latexsym}
\usepackage[T1]{fontenc}
\usepackage[utf8]{inputenc}
\usepackage{microtype}
\usepackage{times}
\usepackage{url}
\usepackage{graphicx,scalerel}
\usepackage{tabularx}
\usepackage{makecell}
\usepackage{multirow}
\usepackage{enumitem}

\usepackage[ruled,vlined]{algorithm2e}
\SetKwRepeat{Do}{do}{while}
\SetAlgoLined
\SetKwInOut{Input}{Input}
\SetKwInOut{Output}{Output}
\usepackage{amsmath,tabstackengine}
\usepackage{booktabs}
\usepackage{caption}
\usepackage{algpseudocode}
\usepackage{amsmath}
\usepackage{graphics}       
\definecolor{green}{rgb}{0.0, 0.5, 0.0}
\definecolor{seagreen}{rgb}{0.0, 0.5, 0.0}
\definecolor{purple}{rgb}{0.47, 0.32, 0.66}
\usepackage{multirow}
\usepackage{rotating}
\usepackage{amssymb}
\usepackage{pifont}
\usepackage{arydshln}
\definecolor{brightpink}{rgb}{1.0, 0.0, 0.5}
\newcommand{\fahimeh}[1]{\textcolor{black}{#1}}

\newcommand{\fahi}[1]{\textcolor{black}{#1}}
\usepackage{bm}
\def\valpha{{\bm{\alpha}}}

\def\vb{{\bm{b}}}

\def\vg{{\bm{g}}}

\def\vx{{\bm{x}}}
\def\vy{{\bm{y}}}

\newcommand{\softmax}{\mathrm{softmax}}

\newcommand{\append}{\mathrm{append}}

\title{Multilingual Neural Machine Translation: \\ Can Linguistic Hierarchies Help?}

   \author{Fahimeh Saleh \quad Wray Buntine \quad  Gholamreza Haffari \quad Lan Du\thanks{~~Corresponding author} \\
  first.last@monash.edu \\ 
  Monash University}

\begin{document}
\maketitle
\begin{abstract}
Multilingual Neural Machine Translation (MNMT) trains a single NMT model that supports translation between multiple languages, 
rather than training separate models for different languages. 
Learning a single model can enhance the
low-resource translation by leveraging data from multiple languages.
However, the performance of an MNMT model is highly dependent on the type of languages used in training,
as transferring knowledge from a diverse set of languages degrades the translation performance due to negative transfer.
In this paper, we propose a Hierarchical Knowledge Distillation (HKD) approach for MNMT which capitalises on language groups generated according to typological features and phylogeny of languages to overcome the issue of negative transfer.
HKD generates a set of multilingual teacher-assistant models via a selective knowledge distillation mechanism based on the language groups, and then distills the ultimate multilingual model from those
assistants in an adaptive way.
Experimental results derived from the TED dataset with 53 languages demonstrate the effectiveness of our approach in avoiding the negative transfer effect in MNMT, leading to an improved translation performance (about 1 BLEU score on average) compared to strong baselines. 
\end{abstract}
\section{Introduction} \label{sec:intro}
The surge over the past few decades in the number of languages used in electronic texts for international communications has promoted Machine Translation (MT) systems to shift towards multilingualism.
However, most successful MT applications, i.e., Neural Machine Translation (NMT) systems, usually rely on supervised deep learning, which is notoriously data-hungry \cite{koehn2017six}.
Despite decades of research, high-quality annotated MT resources are only available for a subset of the world's thousands of languages \fahimeh{\cite{paolillo2006evaluating}}.
Hence, data scarcity is one of the significant challenges which comes along with the language diversity and multilingualism in MT.
One of the most widely-researched approaches to tackle this problem is unsupervised learning which takes advantage of available unlabeled data in multiple languages \cite{lample2017unsupervised,arivazhagan2019missing,snyder2010unsupervised,xu2019polygon}. 
However, unsupervised approaches have relatively lower performance compared to their supervised counterparts \cite{dabre2020survey}.
Nevertheless, the performance of the supervised MNMT models is highly dependent on the types of languages used to train the model \cite{tan2019multilingual1}.  
If languages are from very distant language families, they can lead to \textit{negative transfer} \cite{torrey2010transfer,rosenstein2005transfer}, causing lower translation quality compared to the individual bilingual counterparts. 

To address this problem, some improvements have been achieved recently with solutions that employ some sort of supervision to guide MNMT using \textit{linguistic typology} \cite{oncevay2020bridging,chowdhury2020understanding,kudugunta2019investigating,bjerva2019language}. 
The linguistic typology provides this supervision by treating the world's languages based on their functional and structural characteristics \cite{linguistic2016survey}.
Taking advantage of this property, which explains both language similarity and language diversity, we aim in our approach to combine two solutions for training an MNMT model: (a) creating a universal, language-independent MNMT model \fahimeh{\cite{johnson2017google};
(b) systematically designing the possible variations of language-dependent MNMT models based on the language relations \cite{maimaiti2019multi}.}

Our approach to preventing negative transfer in MNMT is to group models which behave similarly in separate language clusters.
Then, we perform a Knowledge Distillation (KD) \cite{darkHinton15} approach by selectively distilling the bilingual teacher models' knowledge in the same language cluster to a multilingual teacher-assistant model. The intermediate teacher-assistant models are representative of their own language cluster. We further adaptively distill knowledge from the multilingual teacher-assistant models to the ultimate multilingual student. 
In summary, 
our main contributions are as follows:  
\begin{itemize}[leftmargin=*,noitemsep, nosep]
    \item We use \textit{cluster-based teachers} in a hierarchical knowledge distillation approach to prevent negative transfer in MNMT. Different from the previous cluster-based approaches in multilingual settings \cite{oncevay2020bridging,tan2019multilingual1},
    our approach makes use of all the clusters with a \textit{universal} MNMT model while retaining the language relatedness structure in a hierarchy.
    \item We distill the ultimate MNMT model from multilingual teacher-assistant models, each of which represents one language family and usually perform better than the individual bilingual models from the same language family. Thus, the  cluster-based teacher-assistant models can lead to a better knowledge distillation compared to a diverse set of bilingual teacher models as used in multilingual KD  \cite{tan2019multilingual2}.
    \item  We explore \textit{a mixture of linguistic features} by utilizing different clustering approaches to obtain the cluster-based teacher-assistants. As the language groups created by different language feature vectors can contribute differently
    to translation, we adaptively distill knowledge from teacher-assistant models to the ultimate student 
    to improve the knowledge gap of the student.
    \item We perform extensive experiments on 53 languages, showing the effectiveness of our approach in avoiding negative transfer in MNMT, leading to an improved translation performance \fahimeh{(about 1 BLEU score on average)} compared to strong baselines. \fahimeh{We also conduct comprehensive ablation studies and analysis, demonstrating the impact of language clustering in MNMT for different language families and in different resource-size scenarios.}
\end{itemize}
\section{Related Work}\label{sec:related-work}
The majority of works on MNMT mainly focus on different architectural choices varying in the degree of parameter sharing in the multilingual setting. 
For example, the works based on the idea of minimal parameter sharing  
share either encoder, decoder, or attention module \cite{firat2017multi,lu2018neural},
and those with complete parameter sharing tend to share entire models \cite{johnson2017google,ha2toward}. 
In general, these techniques 
implicitly assume that a set of languages is pre-given without considering the positive or negative effect of language transfer 
between the languages shared in one model. 
Hence, they can usually achieve comparable results with individual models (trained with individual language pairs) only when the languages are less diverse or the number of languages is small. 
When several diverse language pairs are involved in training an MNMT system, 
the \textit{negative transfer} \cite{torrey2010transfer,rosenstein2005transfer} usually happens between more distant languages, resulting in degraded translation accuracy in the multilingual setting. 
To address this problem, \citet{tan2019multilingual1}  suggested a clustering approach using either prior knowledge of language families or using language embedding. 
They obtained the language embedding by retrieving the representation of a language tag which is added to the input of an encoder in a universal MNMT model.
Later,  \citet{oncevay2020bridging} introduced another clustering technique using the multi-view language representation.
They fused language embeddings learned in an MNMT model with
syntactic features of a linguistic knowledge base \cite{WALS}.  \citet{tan2019multilingual2} proposed a knowledge distillation approach which transfers knowledge from bilingual teachers to a multilingual student when the accuracy of teachers are  higher than the student. 
Their approach eliminates the accuracy gap between the bilingual and multilingual NMT models.
%
However, we argue that distilling knowledge from a \emph{diverse} set of parent models into a student model can be sub-optimal, as the parents may compete instead of collaborating with each other,
resulting in negative transfer. 
\section{Hierarchical Knowledge Distillation}
\label{sec:hkd}
We address the problem of \textit{data scarcity} and \textit{negative transfer} in MNMT with a \emph{Hierarchical Knowledge Distillation} (HKD) approach. 
\begin{figure*}[!t]
\centering
\includegraphics[width=0.7\textwidth]{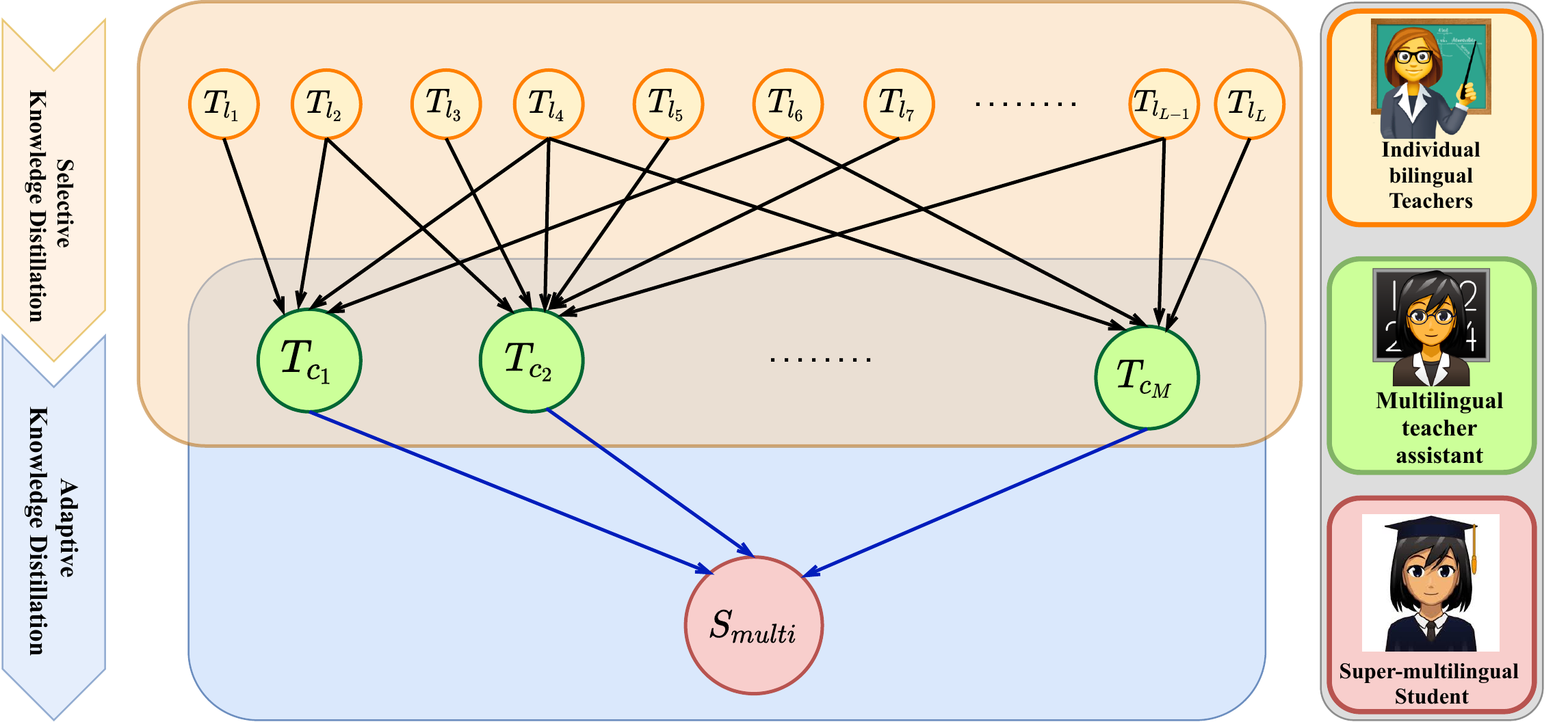}
\vspace{-1.5mm}
\caption{\textbf{HKD} approach: In the first phase of knowledge distillation, aka ``Selective KD'', the knowledge is transferred from bilingual teacher models per clusters (orange circles) to the multilingual teacher-assistant models (green circles). For example $T_{l_1}$, $T_{l_2}$, $T_{l_4}$, and $T_{l_6}$ are belonged to one cluster and distilled to teacher-assistant model ${T}_{c_1}$. In the second KD phase, aka ``Adaptive KD'', knowledge is transferred from ensemble of intermediate related teacher-assistant models to the ultimate student (red circle) adaptively.}
\vspace{-3mm}
\label{fig:HKD}
\end{figure*}
The hierarchy in HKD is constructed in such a way that 
the node structure captures the similarity structure and the relatedness of the languages. 
Specifically, in an inverse pyramidal structure as shown in Figure~\ref{fig:HKD},
the root node corresponds to the ultimate MNMT model that we aim to train,
the leave nodes correspond to each individual bilingual NMT models,
and the non-terminal nodes represent the language clusters.
Our hypothesis is that 
leveraging common characteristics of languages in the same language group, which is formed using clustering algorithms based on the typological properties of languages \cite{linguistic2016survey},
the HKD method can train a high quality MNMT model by
distilling knowledge from related languages, rather than diverse ones.


Our HKD approach consists of two knowledge distillation mechanisms, providing two levels of supervision for training the ultimate MNMT model (illustrated in Figure~\ref{fig:HKD}), including: (i) \textbf{selective distillation} of knowledge from individual bilingual teachers to the multilingual intermediate teacher-assistants, each of which corresponds to one language group;
and (ii) \textbf{adaptive distillation} of knowledge from all related cluster-wise teacher-assistants to the super-multilingual ultimate student model in each mini-batch of training per language pair.
\fahimeh{Note that we do not utilize multilingual adaptive KD in both distillation phases as we need to have the predictions of all the \textit{relevant experts} in adaptive KD. Using adaptive KD for both stages is particularly impractical when there is a huge set of diverse teachers as in the first phase. Hence, in the first distillation phase, we aim to generate the cluster-wise teacher assistants  using selective KD as the pre-requisites for the adaptive KD phase.}
The main steps of HKD are elaborated as follows:

\vspace{5pt}
\noindent
\textbf{Clustering:} 
Clustering can be conducted using 
different language vectors such as: i) sparse language vectors from typological knowledge base (KB) databases, ii) dense learned language embedding vectors from multilingual NLP tasks, and iii) the combination of KB and task-learned language vectors.
The implicit causal relationships between languages are usually learned from translation tasks;
the genetic, the geographical, and the structural similarities between languages are extracted from typlogical KBs \cite{bjerva2019language}. 
Thus, the language groups created by different language vectors can contribute differently
to the translation and it is not quite clear which types of language features are more helpful in MNMT systems \cite{oncevay2020bridging}.
For example, \textit{``Greek''} can be clustered with \textit{``Arabic''} and \textit{``Hebrew''} based on the mix of KB and task-learned language vectors. Meanwhile, it can be clustered with\textit{ ``Macedonian''} and \textit{``Bulgarian''} based on NMT-learned language vectors.
Therefor, we cluster the languages based on all types of language representations and propose to explore a mixture of linguistic features by utilizing all 
clusters in training the ultimate MNMT student. 
So, given a training dataset consisting of $L$ languages and $K$ clustering approaches,
where each clustering approach creates $n$ clusters, 
we are interested in training a many-to-one MNMT model (ultimate student)
by hierarchically distilling knowledge from all $M$ clusters to the ultimate student, where $M := \sum\nolimits_{k=1}^K n_k$.  

\begin{figure}[t]
\centering
\includegraphics[width=0.37\textwidth]{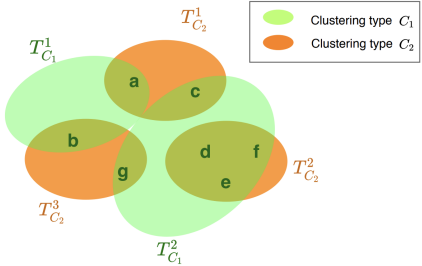}
\vspace{-3mm}
\caption{Effective teachers for each language after clustering. $C$ refers to the clustering type and $T$ refers to the Teacher. For language \textbf{\textit{a}}, we have two \textbf{\textit{effective}} teachers: $T_{C_1}^1$ and $T_{C_2}^1$.}
\vspace{-10pt}
\label{fig:merge-clustering}
\end{figure}

\vspace{5pt}
\noindent
\textbf{Multilingual selective knowledge distillation:} 
Assume we have a language cluster that 
consists of $L^{\prime}$ languages, 
where
$l \in \{1, 2, \dots, L^{\prime}\}$.
Given a collection of pretrained individual teacher models $\{\theta^l\}_{l=1}^{L^{\prime}}$, each handling one language pair in $\{\mathcal{D}^l\}_{l=1}^{L^{\prime}}$, 
and inspired by \citet{tan2019multilingual2}, we use the following knowledge distillation objective for each language $l$ in the cluster.
\begin{equation}
\begin{split}
    \mathcal{L}_{KD}^{selective} ( \mathcal{D}^l, \theta^{c},  \theta^l) := -   \sum\nolimits_{\vx,\vy \in \mathcal{D}^l} \sum\nolimits_{t=1}^{|\vy|} \\ 
   \hspace{-10pt}\sum\nolimits_{v \in V} Q(v | \vy_{<t} , \vx , \theta^l) \log P(v | \vy_{<t} , \vx , \theta^{c}) 
    \label{skd_loss_1}
    \end{split}
\end{equation}
where $\theta^{c}$ is the teacher assistant model, $|V|$ is the vocabulary set, $P(\cdot \mid \cdot)$ is the conditional probability of the teacher assistant model, and $Q( \cdot \mid \cdot)$ denotes the output distribution of the bilingual teacher model.  
According to Eq. (\ref{skd_loss_1}), knowledge distillation regularises the predictive probabilities generated by a cluster-wise multilingual model with those generated by each individual bilingual models.
Together with the translation loss ($\mathcal{L}_{NLL}$), 
we have the following selective KD loss to generate the intermediate teacher-assistant model:
\begin{eqnarray}
\lefteqn{\mathcal{L}_{ALL}^{selective} ( \mathcal{D}^l, \theta^{c},  \theta^l) := } &&  \label{skd_loss_2} \\
&&\hspace*{-15pt}{(1-\lambda)} \mathcal{L}_{NLL} ( \mathcal{D}^l, \theta^{c}) +  {\lambda} \mathcal{L}_{KD}^{selective} ( \mathcal{D}^l, \theta^{c},  \theta^l) \nonumber
\end{eqnarray}
where ${\lambda}$ is a tuning parameter that balances the contribution of the two losses. 
Instead of using all language pairs in Eq~\eqref{skd_loss_1},
we used a deterministic but dynamic approach to 
exclude language pairs from the loss function
if the multilingual student surpasses the individual models on some language pairs during the training, which makes the training selective.
\begin{algorithm}[t]
\scriptsize
\Input{ Training corpora: $\{\mathcal{D}^l\}_{l=1}^L$, where $\mathcal {D}^l:=\{(\vx_1^l,\vy_1),..,(\vx_n^l,\vy_n)\}$ ;\\
List of languages:$L$;\\
List of language clusters: $\{C^{m}\}_{m=1}^M$ ;\\
Cluster-based MNMT models: $\{\theta^c\}_{c=1}^{M}$ ; \\
Total training epochs: $N$; \\
Batch size: J;}
\Output{Ultimate multilingual student model: $\theta_{s}$;\\}
Randomly initialize multilingual model $\theta_{s}$, accumulated gradient $g=0$, distillation flag $f^l=True$ for $l \in L$ \;  
$n = 0$ \;
\While{$n < N$}{
    $g=0$\;
    $C_{sim}=[ ]$\;
    \For{$l \in L$}{
    \textcolor{gray}{ // find the effective clusters with similar languages\;}
    \For{$c \in \{C\}_{1}^M $}{
        \If{$ l \in c $}{
            $C_{sim}.\append(c)$}
    }
    $D^{l} = random\_permute(\mathcal{D}^{l})$ \;
     $\vb_1^l,..,\vb_J^l = create\_minibatches(\mathcal{D}^{l})$ \\ \textcolor{gray}{ //where $b^l=(x^l,y)$  \;}
        $j = 1$ \;
    \While{$j \le J$}{
    \textcolor{gray}{ // compute contribution weights\;}
      \For{$c \in C_{sim}$}{
        $\Delta_{c} = - ppl(\theta^c(b_j^l))$ \;
          }
     $\valpha = \softmax (\Delta_{1},..,\Delta_c)$ \; 
     \textcolor{gray}{ //compute the gradient on loss $\mathcal{L}_{ALL}^{adapt.}$\;}
     $\vg = \nabla_{\theta_{s}} \mathcal{L}_{ALL}^{adapt.}(\vb_j^l,\theta_{s},\{\theta^{c}\}_1^C, \valpha)$  \;
     \textcolor{gray}{ // updates the parameters \;}
      $\theta_{s} = \textrm{update\_param}(\theta_{s},\vg) $ \;
       $j = j + 1$ \; } 
     }
     $n = n + 1$ \;
}
\caption{\small Multilingual Adaptive KD}
\label{ref:algorith-multi-adaptive-kd}
\end{algorithm}
This selective distillation process\footnote{
The training algorithm of selective knowledge distillation is summarized in Alg. 1 in (section A.1 - Appendix),
which is similar to the one used in \cite{tan2019multilingual2}.
} is applied to all clusters obtained from different clustering approaches.
It is noteworthy that (i) the selective knowledge distillation generates
a teacher-assistant model for each cluster, i.e., $c \in \{1,2, \dots, M\}$;
(ii) each language can be in multiple clusters due to the use of different language representations, 
thus there can be
more than one effective teacher-assistant model for any given language pair (illustrated in Figure~\ref{fig:merge-clustering}.). \fahimeh{So for each language pair, we have a set of effective clusters: $c \in \{1,2, \dots, C_{sim}\}$}.

\vspace{5pt}
\noindent
\textbf{Multilingual adaptive knowledge distillation:} Given a collection of effective teacher-assistant models $\{\theta^c\}_{c=1}^{C_{sim}}$, where ${C_{sim}}$ is the number of effective clusters per language, 
we devise the following KD objective for each language pair,
\begin{equation}
\begin{split}
\mathcal{L}_{KD}^{adaptive} (\mathcal{D}^l, \theta_{s},  \{\theta^c\}_{1}^{{C_{sim}}},\valpha) :=  -\sum_{c=1 }^{{C_{sim}}} \sum_{\vx,\vy \in \mathcal{D}^l} \\ \valpha_{c} \sum_{t=1}^{|\vy|} \sum_{v \in V} Q(v | \vy_{<t} , \vx , \theta^c)
  \log P(v | \vy_{<t} , \vx , \theta_{s})
\end{split}
\end{equation}
where $\valpha$ dynamically weigh the contribution of the teacher-assistants/clusters.
$\valpha$ is computed via an attention mechanism
based on the rewards (negative perplexity) attained by the
teachers on the data, where these values are passed through a softmax transformation to turn into a distribution \cite{saleh2020collective}. 
\begin{table*}[!htbp]
    \centering
    \small
    \scalebox{0.63}{
    \begin{tabular}{c | c | c| c | c | c | c | c | c | c }
        \hline
        \thead{cluster 1\\}  & cluster 2  & cluster 3 & cluster 4 & cluster 5 & cluster 6 & cluster 7 & cluster 8 & cluster 9 & cluster 10\\
        \hline
         \thead{Japanese\\ Korean\\ Mongolian \\Burmese \\ \\} & \thead{Malay\\Thai} \thead{ Chinese \\ Indonesian\\Vietnamese} &
         \thead{Marathi \\ Tamli \\ Bengali \\ Georgian \\ \\} & 
         \thead{Kurdish\\ Persian \\ Kazakh} \thead{Basque \\ Hindi \\ Urdu} & 
         \thead{ Greek \\ Arabic \\ Hebrew }& 
         \thead{Turkish \\ Azerbaijani \\ Finnish } \thead{Hungarian \\ Armenian} & 
         \thead{Slovak\\Polish \\ Russian} \thead{ Macedonian \\Lithuanian \\ Belarusian \\Ukrainian} &
         \thead{Croatian \\ Bosnian \\Slovenian\\ Serbian}\thead{Czech \\ Estonian \\ Albanian}&
         \thead{Galician \\ Italian \\ Portuguese}\thead{Bulgarian \\ Romanian \\ Spanish} &
         \thead{French\\ Danish\\Swedish} \thead{Dutch\\German\\Bokmal}\\ 
        \hline
    \end{tabular}}
    \vspace{-2.5mm}
    \caption{\textbf{Clustering type (1) }-- SVCCA-53 \protect\cite{oncevay2020bridging}, Based on  multi-view representation using both syntax features  of WALS and language vectors learned by MNMT model trained with TED-53.}\label{tab:clus1}
\end{table*}
\begin{table*}[!htbp]
\vspace{-5pt}
    \centering
    \small
    \scalebox{0.61}{
    \begin{tabular}{c | c | c| c | c | c | c | c | c | c }
        \hline
        \thead{cluster 1\\}  & cluster 2  & cluster 3 & cluster 4 & cluster 5 & cluster 6 & cluster 7 & cluster 8 & cluster 9 & cluster 10\\
        \hline
         \thead{Korean\\ Bengali \\ Marathi} \thead{Hindi \\ Urdu}& \thead{Basque\\Arabic\\Hebrew} &
         \thead{Armenian \\ Persian \\ Kurdish}&
         \thead{Hungarian \\ Azerbaijani \\ Mongolian} \thead{Turkish\\Japanese}& 
         \thead{Georgian \\ Tamli}&
         \thead{Kazakh\\ Burmese} &
         \thead{Bosnian \\ Albanian \\ Polish \\ Slovak} \thead{Croatian \\ Macedonian \\ Belarusian \\ Estonian} &
         \thead{Russian \\ Ukrainian \\ Slovenian \\ Lithuanian} \thead{Finnish \\ Czech \\Serbian}&
         \thead{ Thai \\ Malay} \thead{ Indonesian \\ Vietnamese\\Chinese}& \thead{Romanian \\ Spanish \\ Italian \\ Galician} \thead{Bulgarian \\ Swedish \\ German } \thead{Dutch \\ Portuguese\\ French} \thead{Danish\\Greek\\Bokmal}\\
        \hline
    \end{tabular}}
    \vspace{-2.5mm}
    \caption{\textbf{Clustering type (2)} -- SVCCA-23 \protect\cite{oncevay2020bridging}, Based on multi-view representation using both syntax features of WALS and language vectors learned by MNMT model trained with WIT-23.}\label{tab:clus2}
\end{table*}
\begin{table*}[!htbp]
\vspace{-5pt}
    \centering
    \small
    \scalebox{0.67}{
    \begin{tabular}{c | c | c| c | c | c | c | c | c | c | c}
        \hline
        \thead{cluster 1\\}  & cluster 2  & cluster 3 & cluster 4 & cluster 5 & cluster 6 & cluster 7 & cluster 8 & cluster 9 & cluster 10 & cluster 11\\
        \hline
         \thead{Estonian \\ Finnish} &
         \thead{Hindi \\ Burmese \\ Armenian \\ Georgian } &
         \thead{Basque\\Bengali  \\ Kurdish \\ Bosnian } \thead{ Belarusian\\ Azerbaijani \\ Mongolian  \\ Marathi} \thead{Urdu\\Tamli\\Kazakh\\Malay}&
         \thead{Galician \\ French \\ Italian \\ Spanish \\ Portuguese} & 
         \thead{Bokmal \\ Danish \\ Swedish}&
         \thead{German \\ Dutch } &
         \thead{Chinese \\ Japanese \\ Korean \\ Hungarian \\ Turkish} &
         \thead{Slovak \\ Polish \\ Russian \\ Ukrainian} \thead{Lithuanian \\ Slovenian \\ Croatian \\ Serbian \\ Czech} &
         \thead{Persian \\ Indonesian \\ Hebrew \\ Vietnamese} \thead{Thai\\Arabic}& 
         \thead{Romanian \\ Albanian} & 
         \thead{Macedonian \\ Bulgarian \\ Greek }\\ 
        \hline
    \end{tabular}}
    \vspace{-2.5mm}
    \caption{\textbf{Clustering type (3)} -- Based on NMT-learned representation using a set of 53 factored language embeddings \protect\cite{oncevay2020bridging, tan2019multilingual1}.}\label{tab:clus3}
\end{table*}
\begin{table*}[!htbp]
\vspace{-5pt}
    \centering
    \small
    \scalebox{0.66}{
    \begin{tabular}{c | c | c}
        \hline
        \thead{cluster 1\\}  & cluster 2  & cluster 3 \\
        \hline
         \thead{ Bengali \\ Chinese \\ Korean} \thead{Kazakh \\ Azerbaijani\\Mongolian} \thead{Burmese\\Japanese\\Marathi} \thead{Urdu\\ Hindi} \thead{Turkish\\Tamli}&
         \thead{Thai \\ Vietnamese \\ Indonesian \\ Malay} &
         \thead{Armenian \\ Georgian \\Macedonian} \thead{Bosnian \\ Slovenian \\ Hungarian} \thead{Basque \\ Serbian\\Dutch} \thead{ Albanian \\ Greek\\ German } \thead{ Galician\\ Romanian\\ Portuguese} \thead{ Croatian\\ Bokmal\\Spanish} \thead{French\\Belarusian\\Lithuanian} \thead{Danish\\Ukrainian\\Hebrew} \thead{Polish\\ Czech \\Russian} \thead{Finnish \\Bulgarian \\Kurdish} \thead{Italian\\Slovak\\ Estonian } \thead{ Arabic\\Persian\\ Swedish}\\ 
        \hline
    \end{tabular}}
    \vspace{-2.5mm}
    \caption{\textbf{Clustering type (4)} -- Based on KB representation using syntax features of WALS \protect\cite{oncevay2020bridging}.}\label{tab:clus4}
\end{table*}
\begin{table*}[t]
    \small
     \centering
    \scalebox{0.6}{
    \begin{tabular}{|c|c|c|c|c|c|c|c|c||||c|c|c|c|c|c|c|c|c|c|c|}
    \hline
        \rotatebox{50}{\textbf{IE/Balto-Slavic}} & \rotatebox{50}{\textbf{IE/Italic}} & 
        \rotatebox{50}{\textbf{IE/Indo-Iranian}} &
        \rotatebox{50}{\textbf{IE/Germanic}} &
        \rotatebox{50}{\textbf{Turkic}} &
        \rotatebox{50}{\textbf{Uralic}} &
        \rotatebox{50}{\textbf{Afroasiatic}} &
        \rotatebox{50}{\textbf{Sino-Tibetan}} &
        \rotatebox{50}{\textbf{Austronesian}} &
        \rotatebox{90}{\textbf{Koreanic}} & 
        \rotatebox{90}{\textbf{Japonic}} & 
        \rotatebox{90}{\textbf{Austroasiatic}} & 
        \rotatebox{90}{\textbf{IE/Hellenic }} & 
        \rotatebox{90}{\textbf{Kra-Dai}} & 
        \rotatebox{90}{\textbf{IE/Albanian}} & 
        \rotatebox{90}{\textbf{IE/Armenian}} & 
        \rotatebox{90}{\textbf{Kartvelian}} & 
        \rotatebox{90}{\textbf{Mongolic}} & 
        \rotatebox{90}{\textbf{Dravidian}} & 
        \rotatebox{90}{\textbf{Isolate (\smaller{Basque}) }} \\
        \hline
        \thead{be, bs, sl, mk, lt, sk\\cs, uk, hr, sr, bg, pl, ru}&
        \thead{bn, ur, ku\\ hi, fa, mr}&
        \thead{gl, pt, ro\\fr, es, it}&
        \thead{nb, da \\sv, de, nl}&
        \thead{kk, az, tr}&
        \thead{et, fi, hu}&
        \thead{he, ar}&
        \thead{my, zh}&
        \thead{ms, id}&
        \thead{ko}&
        \thead{ja}&
        \thead{vi}&
        \thead{el}&
        \thead{th}&
        \thead{sq}&
        \thead{hy}&
        \thead{ka}&
        \thead{mn}&
        \thead{ta}&
        \thead{eu}\\ 
        
    \hline
    \end{tabular}
    }
    \caption{Language families \cite{eberhard2019largest}. IE refers to Indo European.}
    \label{tab:lang_family}
\end{table*}
 
This adaptive distillation of knowledge allows the student model to get the best of teacher-assistants (which are representative of different linguistic features) based on their effectiveness to improve the knowledge gap of the student.
The total loss function then becomes a weighted combination of losses coming from the ensemble of teachers and the data,
\begin{eqnarray}
\lefteqn{\mathcal{L}_{ALL}^{adaptive} (\mathcal{D}^l, \theta_{s},  \{\theta^c\}_{1}^{{C_{sim}}},\valpha) := } \\
&&\hspace*{-22pt}{\lambda}_1 \mathcal{L}_{NLL} (\mathcal{D}^l, \theta_{s}) + 
\lambda_2 \mathcal{L}_{KD}^{adapt.} (\mathcal{D}^l, \theta_{s},  \{\theta^c\}_{1}^{{C_{sim}}},\valpha)
\nonumber
\end{eqnarray}
The training process is summarized in Alg. \ref{ref:algorith-multi-adaptive-kd}. 

\section{Experiment}
In this section we explain our experiment settings as well as our experimental findings.

\noindent
\textbf{Data}: We conducted extensive experiments on a parallel corpus (53 languages$\rightarrow$English) from TED talks transcripts
\footnote{https://github.com/neulab/word-embeddings-for-nmt}
created and tokenized by \citet{qi2018and}. 
This corpus has 26\% of language pairs having less than or equal to 10k sentences (extremely low-resource), and 33\% of language pairs having less than 20k sentences (low-resource).
All the sentences were segmented with BPE segmentation \cite{sennrich2015neural} \fahi{with a learned BPE model with 32k merge operations on all languages.} 
We kept the output vocabulary of the teacher and student models the same to make the knowledge distillation feasible. Details about the size of training data, language codes, and preprocessing steps are described in Section A.2.

\noindent
\textbf{Clustering}: 
We clustered all the languages based on the three different types of representations discussed in Section \ref{sec:hkd}
in order to take advantage of a mixture of linguistic features while training the ultimate student. 
Following \citet{oncevay2020bridging}, we adopted 
their multi-view language representation approach that uses Singular Vector Canonical Correlation Analysis -- SVCCA \cite{raghu2017svcca} to fuse
the one-hot encoded KB representation obtained from syntactic features of WALS \cite{WALS} and a dense NMT-learned view obtained from MNMT \cite{tan2019multilingual1}. 
Specifically, \textit{SVCCA-53} uses 53 languages of TED dataset to build the language representations and generates 10 clusters, the languages within each of which usually have the same \textit{phylogenetic} or \textit{geographical} features.
\textit{SVCCA-23} instead uses 23 languages of  WIT-23 \cite{cettolo2012wit3} to compute the shared space.
We also generated language clusters based on either KB-based representation using syntax features of WALS \cite{WALS}
and NMT-learned representation alone. Tables \ref{tab:clus1}-\ref{tab:clus4} show the generated language clusters.
\\
\noindent
\textbf{Training Configuration}:
All models were trained with Transformer architecture \cite{Vaswani:17}.
The individual baseline models were trained with the model hidden size of 256, 
feed-forward hidden size of 1024, and 2 layers. 
All multilingual models were trained with the model hidden size of 512, feed-forward hidden size of 1024, and 6 layers.  We used selective multilingual knowledge distillation approach for training our cluster-based MNMT models as selective multilingual KD perform better than universal MNMT without KD \cite{tan2019multilingual2}.
We did not carry out intense parameter tuning to search for the best parameter settings of each model for the sake of simplicity. 
It is noteworthy that the purpose of our experiments is rather to demonstrate the benefit of considering different type of language clusters through a unified hierarchical knowledge distillation. Details about the training configuration is provided in Section A.2.
\subsection{Experimental Results}
The translation results of (53 languages $\rightarrow$ English) for all approaches
are summarised in Table \ref{tab:result-all-multi}. 
The language pairs are sorted based on the size of training data in an ascending order. The translation quality is evaluated and reported based on the BLEU  score \cite{papineni_bleu:_2002}.
\begin{table}[!t]
    \centering
    \small
    \hspace*{-1pt}\scalebox{0.69}{
    \setlength\tabcolsep{2.5pt}
    \begin{tabular}{|c|c | c || c c ||c |c c c c|| c|}
        \hline \multirow{2}{*}{\rotatebox{90}{\thead{\textbf{ Resource}}}} &
       \multirow{2}{*}{\thead{\textbf{src}} } &  \multirow{2}{*}{\thead{\textbf{size} \\ \#sent.\\(k)} } &\multicolumn{2}{c||}{ \thead{\textbf{Baseline}\\}}& \multicolumn{5}{c||}{\textbf{Multilingual Selective KD}}& \multirow{2}{*}{ \thead{\textbf{ Multi.}\\\textbf{HKD}}}\\
       \cline{3-10}
         &  &   &  \thead{Individ.} & \thead{ Universal \\Multi.} & \thead{All\\langs} & \thead{Clus. \\type1} &\thead{Clus. \\type2}& \thead{Clus. \\type3}& \thead{Clus. \\type4} & \\
       \hline
\multirow{15}{*}{\rotatebox{90}{\textbf{Extremely low resource}}} & & & & & & & & & &\\
& kk & 3.3 & 3.42 & 5.05 & 4.66 & \underline{7.00} & 3.51 & 3.10 & \textbf{8.13} & 6.61 \\
& be & 4.5 & 5.13 & 12.51 & 12.36 & 12.78 & 10.81 & 8.46 & \textbf{15.18} & \underline{14.88}\\
& bn & 4.6 & 5.06 & 12.50 & \underline{12.58} & 9.13 & 10.11 & 12.16 & 10.13 & \textbf{12.81}\\
& eu & 5.1 & 4.4 & \textbf{13.12} & 12.00 & 9.08 & 9.70 & 8.14 & 11.03 &\underline{12.90}\\
& ms & 5.2 & 3.78 & 13.88 & \textbf{14.61 }& 12.93 & 12.94 & 7.63 & 12.98 & \underline{14.11}\\
& bs & 5.6 & 7.92 & 14.82 & 15.46 & \underline{18.05} & 16.89 & 9.03 & \textbf{19.02} & 17.51\\
& az & 5.9 & 5.79 & \underline{10.32} & 9.91 & 9.59 & 9.17 & 8.64 & 9.23 & \textbf{10.40} \\
& ur & 5.9 & 8.98 & 12.76 & \textbf{16.50} & 13.35 & 13.17 & 12.02 & 13.39 & \underline{15.95}\\
& ta & 6.2 & 4.57 & \underline{5.86} & \textbf{6.19} & 4.02 & 5.76 & 3.96 & 3.49 & 5.63\\
& mn & 7.6 & 3.54 & 6.11 & 5.82 & 5.75 & \underline{6.60} & 5.39 & 6.20 & \textbf{6.81}\\
& mr & 9.8 & 6.92 & 10.53 & \underline{10.72} & 8.70 & 9.00 & 8.39 & 9.04 & \textbf{10.98}\\
& gl & 10.0 & 13.5 & 22.04 & 22.44 & 25.53 & 26.81 & \underline{26.93 }& 25.90 & \textbf{27.11}\\
& ku & 10.3 & 6.3 & 10.32 & 12.12 & 9.43 & 6.98 & 9.22 & \underline{12.93} &\textbf{ 13.03}\\
& et & 10.7 & 8.24 & 12.21 & 13.19 & 13.47 & 13.21 & 10.44 & \underline{13.94} & \textbf{14.10}\\
\hline \hline
\multirow{5}{*}{\rotatebox{90}{\textbf{Low resource}}} & & & & & & & & & &\\
& ka & 13.1 & 8.64 & 8.18 & 8.66 & 9.28 & \underline{10.85} & 9.14 & 8.88 & \textbf{11.15}\\
& nb & 15.8 & 26.36 & 28.49 & 29.08 & 33.55 & \textbf{34.31 }& 28.79 & 30.87 & \underline{33.89}\\
& hi & 18.7 & 10.66 & 16.03 & \textbf{17.93} & 13.27 & 12.09 & 12.16 & 12.80 & \underline{16.11}\\
& sl & 19.8 & 11.45 & 15.12 & 15.39 & 16.48 & 16.54 & 17.75 & \underline{18.31 }& \textbf{18.43} \\
& &  & & &  &  & & & & \\
\hline \hline
\multirow{36}{*}{\rotatebox{90}{\textbf{Enough resource}}} & & & & & & & & & & \\
& hy & 21.3 & 11.14 & 14.07 & 15.12 & 13.76 & 12.77 & 10.81 & \textbf{17.17} & \underline{16.72}\\
& my & 21.4 & 4.91 & \underline{10.70} & \textbf{11.11} & 9.65 & 6.35 & 8.48 & 9.54 & 8.81\\
& fi & 24.2 & 8.16 & 11.69 & 12.23 & 11.36 & 12.57 & 10.59 & \underline{12.76} & \textbf{12.90}\\
& mk & 25.3 & 18.32 & 20.63 & 21.09 & 21.48 & 20.06 & \underline{24.65} & 23.8 & \textbf{25.05}\\
& lt & 41.9 & 14.78 & 15.44 & 16.76 & 16.98 & 16.9 & 17.96 & \textbf{18.24} & \underline{18.11}\\
& sq & 44.4 & 22.62 & 24.44 & 25.22 & 24.74 & 23.22 & \underline{26.89} & 26.42 & \textbf{26.93}\\
& da & 44.9 & 31.85 & 30.39 & 30.61 & 35.02 & \textbf{39.76} & 30.58 & 32.04 &\underline{36.00}\\
& sv & 56.6 & 27.2 & 27.18 & 26.84 & 31.36 & \textbf{34.52} & 26.81 & 28.65 &\underline{33.14}\\
& sk & 61.4 & 19.36 & 22.04 & 22.58 & 22.49 & 21.18 & \underline{24.08} & 23.77 & \textbf{24.33}\\
& id & 87.4 & 20.51 & 20.89 & 20.69 & 21.11 & 21.13 & \underline{22.56} & 21.12 & \textbf{22.76}\\
& th & 96.9 & 20.46 & 21.34 &  21.72 & 22.94 & 22.94 &   \underline{23.09} & 22.87 &  \textbf{23.30}\\
& cs & 103.0 & 20.13 & 22.01 & 22.07 & 21.72 & 22.49 & \underline{23.12} & 22.86 & \textbf{23.62}\\
& uk & 108.4 & 21.32 & 22.11 & 23.07 & 23.06 & 22.91 & 23.58 & \underline{23.66} & \textbf{24.09}\\
& hr & 122.0 & 25.89 & 26.51 & 27.17 & 27.56 & 25.62 & \underline{28.66} & 28.34 & \textbf{28.91} \\
& el & 134.3 & 26.82 & 26.07 & 28.51 & 30.05 & \textbf{31.35} & 29.13 & 29.66 & \underline{30.10} \\
& sr & 136.8 & 26.94 & 25.43 & 25.88 & 27.48 & 25.75 &\underline{ 27.69} & 27.12 & \textbf{27.97}\\
& hu & 147.1 & 18.46 & 17.61 & 18.55 & 19.08 & 18.41 & \textbf{20.16} & 19.82 & \underline{20.10}\\
& fa & 150.8 &\textbf{23.60} & 21.7 & 21.29 & 21.31 & 22.44 & \underline{23.51} & 22.24 & 23.19\\
& de & 167.8 & 15.23 & 14.83 & 16.69 & 16.88 & \underline{17.79} & 15.44 & 16.67 & \textbf{18.04}\\
& ja & 168.2 & 10.11 & 8.61 & 8.93 & 10.14 & 10.14 & \underline{10.17}  & 8.69 & \textbf{10.30}  \\
& vi & 171.9 & 18.97 & 19.19 & 20.58 & 21.60 & 21.60 & \underline{21.30} & 20.33 & \textbf{21.82}\\
& bg & 174.4 & 28.85 & 27.66 & 29.14 & 31.67 & \underline{32.18} & 29.86 & 30.48 & \textbf{32.33}\\
& pl & 176.1 & 17.23 & 18.62 & 19.45 & 19.45 & 18.37 & 19.93 & \underline{20.26} & \textbf{20.71}\\
& ro & 180.4 & 25.21 & 25.97 & 26.53 & 28.03 & 28.43 & \underline{28.90} & 27.35 & \textbf{29.00}\\
& tr & 182.3 & 17.72 & 10.2 & 10.01 & \underline{18.66} & 18.19 & \textbf{19.85} & 18.27 & 16.91 \\
& nl & 183.7 & 27.65 & 26.91 & 26.82 & 28.05 & 29.07 & \underline{29.17} & 28.03 & \textbf{29.58}\\
& zh & 184.8 & 20.44 & 22.10 &  22.71&  22.19 & 22.11 & \textbf{23.91}& 22.84 & \underline{23.85}\\
& es & 195.9 & 30.17 & 29.55 & 30.00 & 29.06 & \textbf{33.45} & 31.46 & 31.82 & \underline{32.76}\\
& it & 204.4 & 26.84 & 25.13 & 27.99 & 30.57 & \underline{30.85} & 30.36 & 29.45 & \textbf{30.93}\\
& ko & 205.4 & 15.98 & 15.71 & 16.41 & 15.17 & 15.18 & \underline{17.45} & 16.00 & \textbf{17.70} \\
& ru & 208.4 & 19.76 & 19.83 & 20.86 & 20.85 & 20.80 & 21.25 & \underline{21.49} & \textbf{21.77}\\
& he & 211.7 & 29.35 & 28.03 & 28.27 & \underline{32.82} & \underline{32.18} & 31.32 & 30.02 & 32.05 \\
& fr & 212.0 & 30.08 & 30.55 & 30.28 & \underline{32.25} & 32.19 & 31.14 & 31.34 & \textbf{32.65}\\
& ar & 213.8 & 25.36 & 23.89 & 24.48 & \underline{28.53} & 28.03 & 27.46 & 25.85 & \textbf{28.94}\\
& pt & 236.4 & 30.99 & 31.12 & 30.85 & 33.36 & \underline{33.84} & 33.25 & 32.56 & \textbf{33.90}\\
& &  & & &  &  & & & & \\
\hline \hline
\textbf{\thead{Avg.\\}}&-&-& 18.50 & 18.64 & 19.24 & 19.84 & 19.87 & 19.35 & \underline{20.05} & \textbf{21.16}  \\
\hline
    \end{tabular}
    }
    \caption{BLEU scores of the translation tasks for 53 Languages$\rightarrow$English. The bold numbers show the best scores and the underline ones show the $2^{nd}$ best results. \fahi{``\textit{All langs}" refers to all 53 languages trained with a multilingual selective KD.}
    }\label{tab:result-all-multi}
\end{table}

\subsubsection{Studies of cluster-based MNMT models}
In this section, we discuss the cluster-based MNMTs' results through the following observations.

\vspace{5pt}
\noindent\textbf{Low resource vs high resource languages}: All cluster-based MNMT and baseline approaches are ranked based on the number of the times they got the first or second-best score in different resource-size scenarios in Table \ref{tab:result-all-clusters-rank}. Based on this result and also the result represented in Table \ref{tab:result-all-multi}, the massive MNMT models with all languages (second column under baseline and first column under selective KD in Tables \ref{tab:result-all-multi}, \ref{tab:result-all-clusters-rank}) outperform the cluster-based MNMT models (columns (2-5) under selective KD in Tables \ref{tab:result-all-multi}, \ref{tab:result-all-clusters-rank}) in extremely low-resource scenarios (e.g., bn-en, ta-en, eu-en). 
This result shows that having more data either from related languages or distant languages has the most impact on training a better MNMT model for under-resourced languages. Furthermore, clustering type (4) is dominant among other clustering approaches for under-resourced situations.
This result is also explainable based on the size of the clusters in clustering type (4).
The translations of extremely low resource languages are significantly improved when they have been clustered in the third cluster of clustering type (4) with 35 languages (shown in Table \ref{tab:clus4}). However, for languages with enough resources, the multilingual baselines with all languages under-performed other cluster-based MNMT models.
\begin{table*}[t]
    \centering
    \scalebox{0.7}{
    \begin{tabular}{c || c || c c ||c |c c c c}
        \hline 
       \multirow{2}{*}{\thead{\textbf{Resource-size} }} &  \multirow{2}{*}{\thead{\textbf{size} (\# sent.)} } &\multicolumn{2}{c||}{ \thead{\textbf{Baseline}\\}}& \multicolumn{5}{c}{\textbf{Multilingual Selective KD}} \\
       \cline{3-9}
         &   &  \thead{Individ.} & \thead{ Multi.} & \thead{All langs} & \thead{Clus. type1} &\thead{Clus. type2}& \thead{Clus. type3}& \thead{Clus. type4} \\
       \hline \hline
       \thead{Extremely low resource} & <= 10k & 0\% & 21.43\% & \textbf{28.57\%} & 14.28\%  & 7.14\% & 3.58\% & \underline{25.00\%} \\ \hline
       \thead{Low resource} & \thead{> 10k and <= 20k} & 0\% & 12.50\% & 12.50\% & \textbf{25.00\%} &\textbf{ 25.00\%} & 12.50\% & 12.50\%\\ \hline
       \thead{Enough resource}  & > 20k & 1.39\% & 1.39\%  & 2.78\% & 20.83\%  & \underline{25.00}\% & \textbf{27.78\%} & 20.83\% \\
        \hline
    \end{tabular}}
    \vspace{-2.5mm}
    \caption{The translation ranking ablation study for all approaches excluding the HKD approach based on the percentage of the times they got the $1^{st}$ or $2^{nd}$ best results. Sum of percentages in each row = 100\%.
    }  
     \label{tab:result-all-clusters-rank}
\end{table*}

\begin{table*}[t]
    \centering
    \scalebox{0.7}{
    \begin{tabular}{c || c || c c ||c |c c c c|| c}
        \hline 
       \multirow{2}{*}{\thead{\textbf{Resource-size}} } &  \multirow{2}{*}{\thead{\textbf{size} (\# sent.)} } &\multicolumn{2}{c||}{ \thead{\textbf{Baseline}}}& \multicolumn{5}{c||}{\textbf{Multilingual Selective KD}}& \multirow{2}{*}{ \thead{\textbf{HKD}}}\\
       \cline{3-9}
         &   &  \thead{Individ.} & \thead{ Multi.} & \thead{All langs} & \thead{Clus. type1} &\thead{Clus. type2}& \thead{Clus. type3}& \thead{Clus. type4} & \\
        \hline \hline
       \thead{Extremely low resource} & <= 10k & 0\% & 13.79\% & 17.24\% & 6.90\% & 3.45\% & 3.45\% & \underline{17.24\%}& \textbf{37.93\%}\\ \hline
       \thead{Low resource} & \thead{> 10k and < 20k} & 0\%  & 0\% & 12.50\% & 0\% & \underline{25.00\%} & 0\% & 12.50\%& \textbf{50.00\%}\\ \hline
       \thead{Enough resource}  & >20k  & 1.43\% & 1.43\%  & 1.43\% & 5.71\% & 12.86\% &\underline{24.28\%} & 8.57\% & \textbf{44.29\%}\\
        \hline
    \end{tabular}}
     \vspace{-2.5mm}
     \caption{The ranking of all approaches based on the percentage of the times they got the $1^{st}$ or $2^{nd}$ best results.
    }  
    \label{tab:result-all-multi-rank}
\end{table*}

\definecolor{blue-random}{rgb}{0.01, 0.28, 1.0}
\begin{table}[t]
\small
\hspace*{-3pt}\scalebox{0.8}{
    \setlength\tabcolsep{1pt}
    \begin{tabular}{|c||c|c||c|c|}
   \hline
       \textbf{\thead{Model\\}}  & \textbf{Contrib. Langs} & \textbf{BLEU} &  \textbf{Contrib. Langs} & \textbf{BLEU}\\
      \hline
       \hline
     \textbf{ \thead{Clus.\\ Rand.1\\} } & \textcolor{blue-random}{\textbf{gl}}, nb, uk, hr, se, ja & 16.53 & \textcolor{blue-random}{\textbf{el}}, id, be & 28.01\\\hline
     \textbf{ \thead{Clus.\\ Rand.2\\}} &  \thead{\textcolor{blue-random}{\textbf{gl}}, ta, mk, be, id, sq, pt, \\fr, ur, az, ku, bs, fa } & 20.27 & \thead{\textcolor{blue-random}{\textbf{el}}, cs, lt, id, sk, th, it,\\ hy, ms, hu, mk, my, bn } & 27.78\\\hdashline
     \textbf{ \thead{Clus.\\ Rand.3\\}} & \textcolor{blue-random}{\textbf{gl}}, az, ja, nb, kk & 13.61& \textcolor{blue-random}{\textbf{el}}, sq, th& 27.97\\\hline
     \textbf{ \thead{Clus.\\ Rand.4\\}} & \thead{\textcolor{blue-random}{\textbf{gl}}, zh, pt, fa, ar, kk, sr,\\ bg, nl, cs, th, ko, vi, hu,\\ mk, fi, ru, mn, de, sl, el,\\ ka, pl, et, ta, fr, ur, ro,\\ sv, mr, be, bs, uk, sq, az}  &22.20 & \thead{\textcolor{blue-random}{\textbf{el}}, zh, pt, fa, ar, kk, sr,\\ bg, nl, cs,th, ko, vi, hu,\\ mk, fi, ru, mn, de, sl, gl,\\ ka, pl, et, ta, fr, ur, ro,\\ sv, mr, be, bs, uk, sq, az} & 28.82\\\hline
      \thead{ \textbf{Avg.}\\}  & - & ${\textbf{18.15}}$ 
      & - &  ${\textbf{28.14}}$ 
      \\\hline
    \end{tabular}}
     \vspace{-2.5mm}
    \caption{Ablation study on using random languages in all clustersing types for (gl$\rightarrow$en) and (el$\rightarrow$en). The results for actual clustering are provided in Table \protect\ref{tab:result-all-multi}. The average result of actual clustering for (gl$\rightarrow$en) and (el$\rightarrow$en) are 26.29 and 30.04 respectively. 
    }
    \label{tab:gl-en}
\end{table}

\noindent
\textbf{Related vs isolated languages}: A group of languages that originated from a similar ancestor is known as a \textit{language family}; and a language that does not have any relationship with another languages is called a \textit{language isolate}. The language families \cite{eberhard2019largest} are shown in Table \ref{tab:lang_family}.
According the results shown in Table \ref{tab:result-all-multi}, clustering approaches usually have the same behaviour and less diversity for clustering languages belonged to IE/Germanic, IE/Italic, Afroasiatic, and Austronesian families. In comparison, there is more diversity, and less consensus for clustering languages belonged to IE/Balto-Slavic, IE/Indo-Iranian, Turkic, Uralic, and Sino-Tibetan families. Moreover, the isolated languages (shown in the last 11 columns of Table \ref{tab:lang_family}) generally have the same behaviour and less variance in BLEU scores in different clustering approaches. This observation shows that cluster-based MNMT models (regardless of the clustering type) do not significantly improve the translation of isolated languages unless the isolated languages have extremely low resources and have been clustered in a huge cluster (e.g., eu-en, hy-en). The results of cluster-based MNMT are presented based on the language families in Section A.3.

\noindent
\textbf{Random clustering vs Actual  clustering:} We conducted an ablation study by using clusters with randomly chosen languages for two translation tasks (el$\rightarrow$en and gl$\rightarrow$en). We kept the number of languages per cluster the same as the actual clustering to make a fair comparison. According to the result represented in Table \ref{tab:gl-en}, for both translation tasks, random clusters underperform the actual clusters in all clustering types. However, notably,  the average BLEU score's difference between the random and actual clusters for gl$\rightarrow$en is considerably higher than el$\rightarrow$en (gl$\rightarrow$en, $\Delta$ = -8.14 \textbf{vs} el$\rightarrow$en, $\Delta$ = -1.9). This observation is inline with the previous observation that \textit{Greek} (el) is an isolated language categorised in IE/Hellenic family and clustering approaches have less impact on this language due to its lower similarity to most  languages. In comparison, \textit{Galician} (gl) is highly similar to the languages in IE/Italic family and clustering improves the translation of gl$\rightarrow$en remarkably. We compared the translation results between actual cluster-based fake cluster-based multilingual approaches in Section A.3.
\subsubsection{Studies of HKD}
According to the results in Table \ref{tab:result-all-multi}, the HKD approach outperforms massive and cluster-based MNMT models in average by 1.11 BLEU score. We discuss it more  
in the following observations:

\noindent
\textbf{Ranking based on data size}: We ranked all approaches, including HKD, based on the number of times they got the first or second-best score (shown in Table \ref{tab:result-all-multi-rank}). According to this result, the HKD approach in three different situations, i.e. extremely low resource, low resource, and enough resource, has the best rank among other approaches. 
This observation proves that the HKD approach is robust in different \textit{data-size} situations by leveraging the best of both multilingual NMT and language-relatedness guidance in a systematic HKD setting.

\vspace{5pt}
\noindent
\textbf{Clustering consistency impact on HKD}: Based on the result in Table \ref{tab:result-all-multi}, the HKD approach underperforms other multilingual approaches when the clusters are inconsistent, causing \fahimeh{a high variance in teacher-assistants’ results. For example for kk$\rightarrow$en, bs$\rightarrow$en, the variance of the BLEU scores of the teacher-assistant models is 6.92 and 20.81 respectively and HKD underachieved a good result.} This observation shows that, although in the second phase of the HKD approach, the cluster-based teacher-assistants
adaptively contribute to training the ultimate students, still a weak teacher-assistant deteriorates the collaborative teaching process. \fahimeh{One possible solution is excluding the worst teacher-assistant in such heterogeneous situations. } 
\subsubsection{Comparison with other approaches}
%
To highlight the pros and cons of the related baselines (with and without KD), we draw a comparison shown in Table \ref{tab:comparison}. Accordingly, our HKD approach is comparable with other approaches based on the following properties: 

\noindent\textbf{Multilingual translation:} Our approach works in a multilingual setting by sharing resources between high-resource and low-resource languages.
This property not only improves the regularisation of the model by avoiding over-fitting to the limited data of the low-resource languages but also decreases the deployment footprint by leveraging the whole training in a single model instead of having individual models per language \cite{dabre2020survey}.

\noindent
\textbf{Optimal transfer}: In the HKD approach, we have an optimal transfer by transferring knowledge from \textit{all} possible languages \textit{related} to a student in the hierarchical structure \fahimeh{which leads to the best average BLEU score  (21.16) comparing to the other baselines (shown in Table \ref{tab:result-all-multi})}. In the universal multilingual NMT without KD, the language transfer stream is maximized when all languages shared their knowledge in a single model during training; however, it is not an optimal transfer due to the lack of any condition on the the relatedness of languages contributing in the multilingual training. \fahimeh{The related experiments are shown in the second column under the baseline experiments in Table \ref{tab:result-all-multi}. The average BLEU score of this approach is 18.46}. In multilingual selective KD \cite{tan2019multilingual2}, knowledge is distilled from one selected teacher with the same language when training the student multilingually. So, although there is a condition on language relatedness, knowledge transfer is not maximized as the similar languages from the same language family are ignored in the distillation process. \fahimeh{The related results are shown in the first column under multilingual selective KD of Table \ref{tab:result-all-multi}. Accordingly, this approach got the average BLEU score of 19.24 in our experiments}. 
 Adaptive KD \cite{saleh2020collective} is a bilingual approach and also uses a random set of teachers which does not essentially have all the related languages to the student and does not lead to optimal transfer. \fahimeh{We did not perform any experiment on adaptive KD \cite{saleh2020collective} since this is a bilingual approach.}

\noindent
\textbf{Adaptive KD vs Selective KD vs HKD}: All KD-based approaches in our comparison reduce \textit{negative transfer} in different ways. 
In multilingual selective KD \cite{tan2019multilingual2}, the risk of negative transfer is reduced by distilling knowledge from the selected teacher per language in a multilingual setting. In bilingual adaptive KD \cite{saleh2020collective}, the contribution weights of different teachers vary based on their effectiveness to improve the student which prevents the negative transfer in bilingual setting. 
In HKD, 
 the hierarchical grouping based on the language similarity provides a systematic guide to prevent negative transfer as much as possible. This property leads HKD to get the \textit{best} results for 32 language pairs out of total 53 language pairs in our multilingual experiments.
\begin{table}[t]
    \centering
    \label{tab:discussion}
    \scalebox{0.61}{
    \begin{tabular}{| c || c| c |c| c |c | }
        \hline
          & \textbf{\thead{Individual\\NMT}}  & \textbf{\thead{Uniform\\ MNMT}} & \textbf{\thead{Selective KD\\MNMT}} & \textbf{\thead{Adaptive KD\\NMT} } & \textbf{\thead{HKD\\MNMT}}
         \\
        \hline
        \hline
         \textbf{\thead{Multilingual\\}} & \textcolor{red}{\ding{55}}& \textcolor{green}{\ding{52}}& \textcolor{green}{\ding{52}} & \textcolor{red}{\ding{55}}&\textcolor{green}{\ding{52}}
         \\
         \hdashline
         \textbf{\thead{Maximum transfer\\}} & \textcolor{gray}{\ding{108}} & \textcolor{green}{\ding{52}}& \textcolor{red}{\ding{55}}& \textcolor{red}{\ding{55}}&\textcolor{green}{\ding{52}}
         \\
          \hdashline
         \textbf{\thead{KD from\\ multiple languages}} &\textcolor{gray}{\ding{108}} & \textcolor{gray}{\ding{108}}&\textcolor{red}{\ding{55}} & \textcolor{green}{\ding{52}}&\textcolor{green}{\ding{52}}
         \\
          \hdashline
         \textbf{\thead{ Reduced risk of\\ negative transfer}} & \textcolor{gray}{\ding{108}}&\textcolor{red}{\ding{55}} &\textcolor{green}{\ding{52}} &\textcolor{green}{\ding{52}} &\textcolor{green}{\ding{52}}
         \\
        \hline
    \end{tabular}
    }
    \vspace{-2.5mm}
    \caption{Comparing different properties of HKD with: transformer-based individual and multilingual NMT \protect\cite{Vaswani:17}, multilingual selective KD \protect\cite{tan2019multilingual2}, and adaptive KD \protect \cite{saleh2020collective}.
    }
    \label{tab:comparison}
    \vspace{-10pt}
\end{table}

\section{Conclusion}
We presented a Hierarchical Knowledge  Distillation (HKD) approach to mitigate the negative transfer effect in MNMT when having a diverse set of languages in training. We put together all languages which behave similarly in the first phase of distillation process and generated the expert teacher-assistants for each group of languages. As we clustered languages based on four different language representations capturing different linguistic features, we then adaptively distill knowledge from all related teacher-assistant models to the ultimate student in each mini-batch of training per language. Experimental results on 53 languages to English show our approach's effectiveness to reduce negative transfer in MNMT. \fahi{Our proposed approach is generalizable to one-to-many with the same setting as a many-to-one task. The clustering needs to be done in the target language, though. For many-to-many tasks, the hierarchical KD can be effective if the clustering is applied to both source and target languages.  Our intended use, however, is many to one or one to many.}

As the future direction, it is interesting to study an end-to-end HKD approach by adding a backward HKD pass compared to the forward HKD pass described in this paper.
\section*{Acknowledgements}
This work was supported by
the Multi-modal Australian ScienceS Imaging and
Visualisation Environment (MASSIVE) \texttt{(www.massive.org.au)}.
\bibliography{emnlp2021}

\begin{thebibliography}{35}
\expandafter\ifx\csname natexlab\endcsname\relax\def\natexlab#1{#1}\fi

\bibitem[{Arivazhagan et~al.(2019)Arivazhagan, Bapna, Firat, Aharoni, Johnson,
  and Macherey}]{arivazhagan2019missing}
Naveen Arivazhagan, Ankur Bapna, Orhan Firat, Roee Aharoni, Melvin Johnson, and
  Wolfgang Macherey. 2019.
\newblock \href {https://arxiv.org/abs/1903.07091} {The missing ingredient in
  zero-shot neural machine translation}.
\newblock \emph{arXiv preprint arXiv:1903.07091}.

\bibitem[{Bjerva et~al.(2019)Bjerva, {\"O}stling, Veiga, Tiedemann, and
  Augenstein}]{bjerva2019language}
Johannes Bjerva, Robert {\"O}stling, Maria~Han Veiga, J{\"o}rg Tiedemann, and
  Isabelle Augenstein. 2019.
\newblock \href
  {https://www.mitpressjournals.org/doi/full/10.1162/COLI_a_00351} {What do
  language representations really represent?}
\newblock \emph{Computational Linguistics}, 45(2):381--389.

\bibitem[{Bowman et~al.(2016)Bowman, Vilnis, Vinyals, Dai, Jozefowicz, and
  Bengio}]{bowman2015generating}
Samuel Bowman, Luke Vilnis, Oriol Vinyals, Andrew Dai, Rafal Jozefowicz, and
  Samy Bengio. 2016.
\newblock \href {https://www.aclweb.org/anthology/K16-1002.pdf} {Generating
  sentences from a continuous space}.
\newblock In \emph{Proceedings of The 20th SIGNLL Conference on Computational
  Natural Language Learning}, pages 10--21.

\bibitem[{Cettolo et~al.(2012)Cettolo, Girardi, and Federico}]{cettolo2012wit3}
Mauro Cettolo, Christian Girardi, and Marcello Federico. 2012.
\newblock \href {https://wit3.fbk.eu/} {Wit3: Web inventory of transcribed and
  translated talks}.
\newblock In \emph{Conference of european association for machine translation},
  pages 261--268.

\bibitem[{Chowdhury et~al.(2020)Chowdhury, Espa{\~n}a-Bonet, and van
  Genabith}]{chowdhury2020understanding}
Koel~Dutta Chowdhury, Cristina Espa{\~n}a-Bonet, and Josef van Genabith. 2020.
\newblock \href {https://www.aclweb.org/anthology/2020.coling-main.532.pdf}
  {Understanding translationese in multi-view embedding spaces}.
\newblock In \emph{Proceedings of the 28th International Conference on
  Computational Linguistics}, pages 6056--6062.

\bibitem[{Dabre et~al.(2020)Dabre, Chu, and Kunchukuttan}]{dabre2020survey}
Raj Dabre, Chenhui Chu, and Anoop Kunchukuttan. 2020.
\newblock \href {https://dl.acm.org/doi/pdf/10.1145/3406095} {A survey of
  multilingual neural machine translation}.
\newblock \emph{ACM Computing Surveys (CSUR)}, 53(5):1--38.

\bibitem[{Dryer and Haspelmath(2013)}]{WALS}
Matthew~S. Dryer and Martin Haspelmath. 2013.
\newblock \href {https://wals.info/} {The world atlas of language structures}.

\bibitem[{Eberhard et~al.(2019)Eberhard, Simons, and
  Fennig}]{eberhard2019largest}
DM~Eberhard, GF~Simons, and CD~Fennig. 2019.
\newblock \href {https://www.ethnologue.com/} {What are the largest language
  families}.
\newblock \emph{Ethnologue: Languages of the World}.

\bibitem[{Firat et~al.(2017)Firat, Cho, Sankaran, Vural, and
  Bengio}]{firat2017multi}
Orhan Firat, Kyunghyun Cho, Baskaran Sankaran, Fatos T~Yarman Vural, and Yoshua
  Bengio. 2017.
\newblock \href
  {https://www.sciencedirect.com/science/article/abs/pii/S0885230816301097}
  {Multi-way, multilingual neural machine translation}.
\newblock \emph{Computer Speech \& Language}, 45:236--252.

\bibitem[{Ha et~al.(2016)Ha, Niehues, and Waibel}]{ha2toward}
Thanh-Le Ha, Jan Niehues, and Alexander Waibel. 2016.
\newblock \href
  {https://workshop2016.iwslt.org/downloads/IWSLT_2016_paper_5.pdf} {Toward
  multilingual neural machine translation with universal encoder and decoder}.
\newblock \emph{Institute for Anthropomatics and Robotics}, 2(10.12):16.

\bibitem[{Hinton et~al.(2015)Hinton, Vinyals, and Dean}]{darkHinton15}
Geoffrey Hinton, Oriol Vinyals, and Jeffrey Dean. 2015.
\newblock \href {http://arxiv.org/abs/1503.02531} {Distilling the knowledge in
  a neural network}.
\newblock In \emph{NIPS Deep Learning and Representation Learning Workshop}.

\bibitem[{Johnson et~al.(2017)Johnson, Schuster, Le, Krikun, Wu, Chen, Thorat,
  Vi{\'e}gas, Wattenberg, Corrado et~al.}]{johnson2017google}
Melvin Johnson, Mike Schuster, Quoc~V Le, Maxim Krikun, Yonghui Wu, Zhifeng
  Chen, Nikhil Thorat, Fernanda Vi{\'e}gas, Martin Wattenberg, Greg Corrado,
  et~al. 2017.
\newblock \href
  {https://www.mitpressjournals.org/doi/pdfplus/10.1162/tacl_a_00065} {Google's
  multilingual neural machine translation system: Enabling zero-shot
  translation}.
\newblock \emph{Transactions of the Association for Computational Linguistics},
  5:339--351.

\bibitem[{Kingma and Ba(2015)}]{kingma2014adam}
Diederik~P. Kingma and Jimmy Ba. 2015.
\newblock \href {https://arxiv.org/abs/1412.6980} {Adam: {A} method for
  stochastic optimization}.
\newblock In \emph{3rd International Conference on Learning Representations,
  {ICLR} 2015, San Diego, CA, USA, May 7-9, 2015, Conference Track
  Proceedings}.

\bibitem[{Koehn and Knowles(2017)}]{koehn2017six}
Philipp Koehn and Rebecca Knowles. 2017.
\newblock \href {https://www.aclweb.org/anthology/W17-3204.pdf} {Six challenges
  for neural machine translation}.
\newblock \emph{ACL 2017}, page~28.

\bibitem[{Kudugunta et~al.(2019)Kudugunta, Bapna, Caswell, and
  Firat}]{kudugunta2019investigating}
Sneha Kudugunta, Ankur Bapna, Isaac Caswell, and Orhan Firat. 2019.
\newblock \href {https://www.aclweb.org/anthology/D19-1167/} {Investigating
  multilingual nmt representations at scale}.
\newblock In \emph{Proceedings of the 2019 Conference on Empirical Methods in
  Natural Language Processing and the 9th International Joint Conference on
  Natural Language Processing (EMNLP-IJCNLP)}, pages 1565--1575.

\bibitem[{Lample et~al.(2017)Lample, Conneau, Denoyer, and
  Ranzato}]{lample2017unsupervised}
Guillaume Lample, Alexis Conneau, Ludovic Denoyer, and Marc'Aurelio Ranzato.
  2017.
\newblock \href {https://arxiv.org/abs/1711.00043} {Unsupervised machine
  translation using monolingual corpora only}.
\newblock \emph{arXiv preprint arXiv:1711.00043}.

\bibitem[{Lu et~al.(2018)Lu, Keung, Ladhak, Bhardwaj, Zhang, and
  Sun}]{lu2018neural}
Yichao Lu, Phillip Keung, Faisal Ladhak, Vikas Bhardwaj, Shaonan Zhang, and
  Jason Sun. 2018.
\newblock \href {https://www.aclweb.org/anthology/W18-6309.pdf} {A neural
  interlingua for multilingual machine translation}.
\newblock In \emph{Proceedings of the Third Conference on Machine Translation:
  Research Papers}, pages 84--92.

\bibitem[{Maimaiti et~al.(2019)Maimaiti, Liu, Luan, and
  Sun}]{maimaiti2019multi}
Mieradilijiang Maimaiti, Yang Liu, Huanbo Luan, and Maosong Sun. 2019.
\newblock Multi-round transfer learning for low-resource nmt using multiple
  high-resource languages.
\newblock \emph{ACM Transactions on Asian and Low-Resource Language Information
  Processing (TALLIP)}, 18(4):1--26.

\bibitem[{O'Horan et~al.(2016)O'Horan, Berzak, Vuli{\'c}, Reichart, and
  Korhonen}]{linguistic2016survey}
Helen O'Horan, Yevgeni Berzak, Ivan Vuli{\'c}, Roi Reichart, and Anna Korhonen.
  2016.
\newblock \href {https://www.aclweb.org/anthology/C16-1123.pdf} {Survey on the
  use of typological information in natural language processing}.
\newblock In \emph{Proceedings of COLING 2016, the 26th International
  Conference on Computational Linguistics: Technical Papers}, pages 1297--1308.

\bibitem[{Oncevay et~al.(2020)Oncevay, Haddow, and Birch}]{oncevay2020bridging}
Arturo Oncevay, Barry Haddow, and Alexandra Birch. 2020.
\newblock \href {https://www.aclweb.org/anthology/2020.emnlp-main.187.pdf}
  {Bridging linguistic typology and multilingual machine translation with
  multi-view language representations}.
\newblock \emph{arXiv preprint arXiv:2004.14923}.

\bibitem[{Ott et~al.(2019)Ott, Edunov, Baevski, Fan, Gross, Ng, Grangier, and
  Auli}]{ott2019fairseq}
Myle Ott, Sergey Edunov, Alexei Baevski, Angela Fan, Sam Gross, Nathan Ng,
  David Grangier, and Michael Auli. 2019.
\newblock \href {https://www.aclweb.org/anthology/N19-4009.pdf}
  {\textsc{fairseq}: A fast, extensible toolkit for sequence modeling}.
\newblock In \emph{Proceedings of the 2019 Conference of the North American
  Chapter of the Association for Computational Linguistics (Demonstrations)},
  pages 48--53.

\bibitem[{Paolillo and Das(2006)}]{paolillo2006evaluating}
John~C Paolillo and Anupam Das. 2006.
\newblock Evaluating language statistics: The ethnologue and beyond.
\newblock \emph{Contract report for UNESCO Institute for Statistics}.

\bibitem[{Papineni et~al.(2002)Papineni, Roukos, Ward, and
  Zhu}]{papineni_bleu:_2002}
Kishore Papineni, Salim Roukos, Todd Ward, and Wj~Zhu. 2002.
\newblock \href {https://www.aclweb.org/anthology/P02-1040/} {{BLEU}: a method
  for automatic evaluation of machine translation}.
\newblock In \emph{{ACL}}.

\bibitem[{Qi et~al.(2018)Qi, Sachan, Felix, Padmanabhan, and
  Neubig}]{qi2018and}
Ye~Qi, Devendra Sachan, Matthieu Felix, Sarguna Padmanabhan, and Graham Neubig.
  2018.
\newblock \href {https://www.aclweb.org/anthology/N18-2084/?ref=hackernoon.com}
  {When and why are pre-trained word embeddings useful for neural machine
  translation?}
\newblock In \emph{Proc. of the 2018 Conference of the North {A}merican Chapter
  of the Association for Computational Linguistics: Human Language
  Technologies, Volume 2 (Short Papers)}, New Orleans, Louisiana. ACL.

\bibitem[{Raghu et~al.(2017)Raghu, Gilmer, Yosinski, and
  Sohl-Dickstein}]{raghu2017svcca}
Maithra Raghu, Justin Gilmer, Jason Yosinski, and Jascha Sohl-Dickstein. 2017.
\newblock \href
  {http://papers.neurips.cc/paper/7188-svcca-singular-vector-canonical-correlation-analysis-for-deep-learning-dynamics-and-interpretability.pdf}
  {{SVCCA}: singular vector canonical correlation analysis for deep learning
  dynamics and interpretability}.
\newblock In \emph{Proceedings of the 31st International Conference on Neural
  Information Processing Systems}, pages 6078--6087.

\bibitem[{Rosenstein(2005)}]{rosenstein2005transfer}
Michael~T Rosenstein. 2005.
\newblock \href
  {http://socrates.acadiau.ca/courses/comp/dsilver/Share/2005Conf/NIPS2005_ITWS/Website/Papers/ITWS10-RosensteinM05_ITWS.pdf}
  {To transfer or not to transfer}.
\newblock In \emph{NIPS-2005 Workshop on Transfer Learning}, volume 898, pages
  1--4.

\bibitem[{Saleh et~al.(2020)Saleh, Buntine, and Haffari}]{saleh2020collective}
Fahimeh Saleh, Wray Buntine, and Gholamreza Haffari. 2020.
\newblock \href {https://www.aclweb.org/anthology/2020.coling-main.302/}
  {Collective wisdom: Improving low-resource neural machine translation using
  adaptive knowledge distillation}.
\newblock In \emph{Proceedings of the 28th International Conference on
  Computational Linguistics}, pages 3413--3421.

\bibitem[{Sennrich et~al.(2016)Sennrich, Haddow, and
  Birch}]{sennrich2015neural}
Rico Sennrich, Barry Haddow, and Alexandra Birch. 2016.
\newblock \href {https://www.aclweb.org/anthology/P16-1162.pdf} {Neural machine
  translation of rare words with subword units}.
\newblock In \emph{Proc. of the 54th Annual Meeting of the Association for
  Computational Linguistics (Volume 1: Long Papers)}, pages 1715--1725, Berlin,
  Germany. ACL.

\bibitem[{Snyder et~al.(2010)}]{snyder2010unsupervised}
Benjamin Snyder et~al. 2010.
\newblock \href
  {https://dspace.mit.edu/bitstream/handle/1721.1/62455/711173292-MIT.pdf?sequence=2&isAllowed=y}
  {\emph{Unsupervised multilingual learning}}.
\newblock Ph.D. thesis, Massachusetts Institute of Technology.

\bibitem[{Tan et~al.(2019{\natexlab{a}})Tan, Chen, He, Xia, Tao, and
  Liu}]{tan2019multilingual1}
Xu~Tan, Jiale Chen, Di~He, Yingce Xia, QIN Tao, and Tie-Yan Liu.
  2019{\natexlab{a}}.
\newblock \href {https://www.aclweb.org/anthology/D19-1089/} {Multilingual
  neural machine translation with language clustering}.
\newblock In \emph{Proceedings of the 2019 Conference on Empirical Methods in
  Natural Language Processing and the 9th International Joint Conference on
  Natural Language Processing (EMNLP-IJCNLP)}, pages 962--972.

\bibitem[{Tan et~al.(2019{\natexlab{b}})Tan, Ren, He, Qin, Zhao, and
  Liu}]{tan2019multilingual2}
Xu~Tan, Yi~Ren, Di~He, Tao Qin, Zhou Zhao, and Tie-Yan Liu. 2019{\natexlab{b}}.
\newblock \href {https://arxiv.org/abs/1902.10461} {Multilingual neural machine
  translation with knowledge distillation}.
\newblock \emph{arXiv preprint arXiv:1902.10461}.

\bibitem[{Torrey and Shavlik(2010)}]{torrey2010transfer}
Lisa Torrey and Jude Shavlik. 2010.
\newblock \href
  {https://ftp.cs.wisc.edu/machine-learning/shavlik-group/torrey.handbook09.pdf}
  {Transfer learning}.
\newblock In \emph{Handbook of research on machine learning applications and
  trends: algorithms, methods, and techniques}, pages 242--264. IGI global.

\bibitem[{Vaswani et~al.(2017)Vaswani, Shazeer, Parmar, Uszkoreit, Jones,
  Gomez, Kaiser, and Polosukhin}]{Vaswani:17}
Ashish Vaswani, Noam Shazeer, Niki Parmar, Jakob Uszkoreit, Llion Jones,
  Aidan~N Gomez, {\L}ukasz Kaiser, and Illia Polosukhin. 2017.
\newblock \href {https://arxiv.org/abs/1706.03762} {Attention is all you need}.
\newblock In \emph{Advances in Neural Information Processing Systems 30}, pages
  5998--6008. Curran Associates, Inc.

\bibitem[{Wang et~al.(2020)Wang, Tsvetkov, and Neubig}]{wang2020balancing}
Xinyi Wang, Yulia Tsvetkov, and Graham Neubig. 2020.
\newblock \href {https://www.aclweb.org/anthology/2020.acl-main.754/}
  {Balancing training for multilingual neural machine translation}.
\newblock In \emph{Proceedings of the 58th Annual Meeting of the Association
  for Computational Linguistics}, pages 8526--8537.

\bibitem[{Xu et~al.(2019)Xu, Qin, Wang, and Liu}]{xu2019polygon}
Chang Xu, Tao Qin, Gang Wang, and Tie-Yan Liu. 2019.
\newblock \href {https://www.ijcai.org/Proceedings/2019/0739.pdf} {Polygon-net:
  A general framework for jointly boosting multiple unsupervised neural machine
  translation models.}
\newblock In \emph{IJCAI}, pages 5320--5326.

\end{thebibliography}
\bibliographystyle{acl_natbib}

\newpage
\section*{Appendix:}
\label{sec:appendix}
\subsection*{Selective Knowledge Distillation}
This first stage of knowledge distillation is the same as the algorithm which is proposed by \cite{tan2019multilingual2}  and is summarized in Algorithm \ref{alg:algorith-selective-kd}. This process is applied to all clusters obtained from different clustering approaches, $\{C^{m}:=\{l_1, ..., l_{|L^{\prime}|}\}\}_{m=1}^M$ where $M$ refers to the number of clusters
and $L^{\prime}$ refers to the number of languages in  each cluster.
\subsection*{Experiment Settings}\label{sec-ap:data}
\begin{figure*}[!htbp]
\centering
\includegraphics[width=0.9\textwidth]{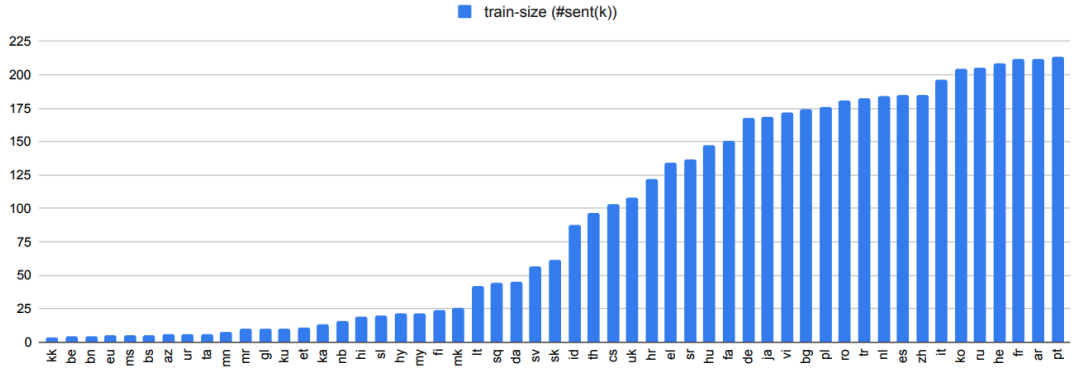}
\caption{Training size (based on the number of sentences) for TED-53 bilingual resources (Language$\rightarrow$English)}
\label{fig:ted-size}
\end{figure*}

\paragraph{Data.} We conducted the experiments on a parallel corpus
from TED talks transcripts\footnote{https://github.com/neulab/word-embeddings-for-nmt} on 53 languages to English created and tokenized by \cite{qi2018and}. Detail about the size of training data and language codes based on ISO 639-1 standard\footnote{http://www.loc.gov/standards/iso639-2/php/English\_list.php} are listed in Table \ref{tab:data-ted-53} and visualised in Figure \ref{fig:ted-size}. We concatenated all data which have the Portuguese-related languages in the source (pt$\rightarrow$en, pt-br$\rightarrow$en). We also concatenated all data with French-related languages in the source (fr$\rightarrow$en, fr-ca$\rightarrow$en). We removed any sentences in the training data which has overlap with any of the test sets. For multilingual training, as a standard practice \cite{wang2020balancing}, 
we up-sampled the data of low-resource language pairs to
make all language pairs having roughly the same size and
adjust the distribution of training data. 
\begin{algorithm}[!t]
\scriptsize
\Input{\small Training corpora: $\{\mathcal{D}^l\}_{l=1}^L$; where $\mathcal {D}^l:=\{(\vx_1^l,\vy_1),..,(\vx_n^l,\vy_n)\}$; \\
List of all languages: $L$;\\
Individual models $\{\theta^l\}_{l=1}^{L^{\prime}}$; \\
List of language pairs per cluster: $L^{\prime}$ ; \\
Total training epochs: $N$;\\
Distillation check step: $\mathcal{N}_{check}$;\\
Threshold of distillation accuracy: \scriptsize $\mathcal{T}$\\}
\vspace{3pt}
\Output{$\theta^{c}$: multilingual model for each cluster,\\}
\vspace{4pt}
Randomly initialize multilingual model $\theta^{c}$, accumulated gradient $g=0$, distillation flag $f^l=True$ for $l \in L^{\prime}$ \; 
$n = 0$ \;
\While{$n < N$}{
    $g=0$\;
    \For{$l \in L^{\prime}$}{
    $D^{l} = random\_permute(\mathcal{D}^{l})$ \;
     $\vb_1^l,..,\vb_J^l = create\_minibatches(\mathcal{D}^{l})$ ,where $b^l=(x^l,y)$ \;
        $j = 1$ \;
    \While{$j \le J$}{
     \If{$f^l == True$}{
     \textcolor{gray}{//compute and accumulate the gradient on loss $\mathcal{L}_{ALL}^{selective}$\;}
     $\vg = \nabla_{\theta^{c}} \mathcal{L}_{ALL}^{selective}(\vb_j^l,\theta^{c},\theta^{l})$  \;
      \textcolor{gray}{// updates the parameters using the optimiser ADAM \;}
      $\theta^{c} = \textrm{update\_param}(\theta^{c},\vg) $ \;} 
      \Else{
      \textcolor{gray}{ //compute and accumulate the gradient on loss $\mathcal{L}_{NLL}$ \;}
     $\vg = \nabla_{\theta^{c}} \mathcal{L}_{NLL}(\vb_j^l,\theta^{c},\theta^{l})$  \;
      \textcolor{gray}{ // updates the parameters using the optimiser ADAM \;}
      $\theta^{c} = \textrm{update\_param}(\theta^{c},\vg) $ \; 
      }
     }
     $j = j + 1$ \; 
     }
     \If{$N\% \mathcal{N}_{check}==0$}{
     \For{$l \in L^{\prime}$}{
     \If{$Accuracy(\theta^c) < Accuracy(\theta^l) + $ \scriptsize $\mathcal{T}$}{
     $f^l=True\;$
     }
     \Else{
     $f^l=False$
     }
     }
    }
    $n = n + 1$ \;
}
\caption{\small Multilingual Selective Knowledge Distillation \protect\cite{tan2019multilingual2}}
\label{alg:algorith-selective-kd}
\end{algorithm}

\begin{table*}[!t]
    \centering
    \scalebox{0.8}{
    \begin{tabular}{cccccccccc}
    \multicolumn{7}{c}{\textbf{TED-53 Languages}}\\& \\
    \bottomrule
    \toprule
        Language name & Kazakh & Belarusian & Bengali & Basque & Malay & Bosnian	\\
     \midrule
       Code &  kk & be & bn & eu & ms & bs\\
     \midrule
     train-size (\#sent(k)) &
     3.3&
     4.5& 
     4.6&
     5.1&
     5.2&
     5.6&\\
\toprule
\toprule
        Language name & Azerbaijani & Urdu & Tamli & Mongolian & Marathi & Galician	\\
     \midrule
       Code &  az & ur & ta & mn & mr & gl\\
     \midrule
     train-size (\#sent(k)) & 
     5.9 &
     5.9& 
     6.2&
     7.6&
     9.8&
     10&\\
\toprule
\toprule    
   Language name & Kurdish & Estonian & Georgian & Bokmal & Hindi & Slovenian	\\
     \midrule
       Code &  ku & et & ka & nb & hi & sl\\
     \midrule
     train-size (\#sent(k)) & 
     10.3 &
     10.7& 
     13.1&
     15.8&
     18.7&
     19.8&\\
\toprule
\toprule    
   Language name & Kurdish & Estonian & Georgian & Bokmal & Hindi & Slovenian	\\
     \midrule
       Code &  ku & et & ka & nb & hi & sl\\
     \midrule
     train-size (\#sent(k)) & 
     10.3 &
     10.7& 
     13.1&
     15.8&
     18.7&
     19.8&\\
\toprule
\toprule    
   Language name & Armenian & Burmese & Finnish & Macedonian & Lithuanian & Albanian	\\
     \midrule
       Code &  hy & my & fi & mk & lt & sq\\
     \midrule
     train-size (\#sent(k)) & 
     21.3 &
     21.4& 
     24.2&
     25.3&
     41.9&
     44.4&\\
\toprule
\toprule    
   Language name & Danish & Swedish & Slovak & Indonesian & Thai & Czech	\\
     \midrule
       Code &  da & sv & sk & id & th & cs\\
     \midrule
     train-size (\#sent(k)) & 
     44.9 &
     56.6& 
     61.4&
     87.4&
     96.9&
     103&\\
\toprule
\toprule    
   Language name & Ukrainian & Croatian & Greek & Serbian & Hungarian & Persian	\\
     \midrule
       Code &  uk & hr & el & sr & hu & fa\\
     \midrule
     train-size (\#sent(k)) & 
     108.4 &
     122& 
     134.3&
     136.8&
     147.1&
     150.8&\\
\toprule
\toprule    
   Language name & German & Japanese & Vietnamese & Bulgarian & Polish & Romanian	\\
     \midrule
       Code &  de & ja & vi & bg & pl & ro\\
     \midrule
     train-size (\#sent(k)) & 
     167.8 &
     168.2& 
     171.9&
     174.4&
     176.1&
     180.4&\\
\toprule
\toprule    
   Language name & Turkish & Dutch & Chinese & Spanish & Italian & Korean 	\\
     \midrule
       Code &  tr & nl & zh & es & it & ko \\
     \midrule
     train-size (\#sent(k)) & 
     182.3 &
     183.7& 
      184.8 &
     195.9&
     204.4&
     205.4&\\
\toprule
\toprule    
   Language name & Russian&  Hebrew & French & Arabic & Portuguese	\\
     \midrule
       Code &  ru &  he & fr & ar & pt \\
     \midrule
     train-size (\#sent(k)) & 
     208.4 &
     211.7 &
     212& 
     213.8&
     236.4&
    \\
     \bottomrule
    \end{tabular}
    }
    \caption{Bilingual resources of 53 Languages $\rightarrow$ English from TED dataset. Language names, language codes based on ISO 639-1 standard\protect\footnote{http://www.loc.gov/standards/iso639-2/php/English\_list.php}, and training size based on the number of sentences in bilingual resources are shown in this table.}
    \label{tab:data-ted-53}
\end{table*}

\paragraph{Training configuration.} All models are trained with Transformer architecture \cite{Vaswani:17},  implemented in the Fairseq framework \cite{ott2019fairseq}. 
The individual models are trained with the model hidden size of 256, feed-forward hidden size of 1024, and 2 layers. All multilingual models either cluster-based or universal MNMT models with or without knowledge distillation were trained with the model hidden size of 512, feed-forward hidden size of 1024, and 6 layers. We use the Adam optimizer \cite{kingma2014adam} and an inverse square root schedule with warmup (maximum LR 0.0005). We apply dropout and label smoothing with a rate of 0.3 and 0.1 for bilingual and multilingual models respectively. 

For the first phase of distillation, i.e., the multilingual selective KD, the distillation coefficient $\lambda$ is equal to 0.6. In the second phase of distillation, i.e., the multilingual adaptive KD, we applied ${\lambda}_1 = 0.5$  and ${\lambda}_2$ is started from 0.5 and increased to 3 using the annealing function of~\cite{bowman2015generating, saleh2020collective}.
We train our final multilingual student with mixed-precision floats on up to 8 V100 GPUs for maximum 100 epochs ($\approx3$ days), with at most 8192 tokens per batch and early stopping at 20 validation steps based on the BLEU score. The translation quality is also evaluated and reported based on the BLEU \cite{papineni_bleu:_2002} score\footnote{SacreBLEU signature:BLEU+case.mixed+numrefs.\\1+smooth.exp+tok.none+version.1.3.1}.
\definecolor{blue-random}{rgb}{0.01, 0.28, 1.0}
\begin{table*}[!t]
\centering
   \hspace*{-2.5pt}\scalebox{0.79}{
    \setlength\tabcolsep{2pt}
    \begin{tabular}{|c||c|c|||c|c|}
   \hline
       \textbf{\thead{Model\\}}  & \textbf{Contributed Langs} & \textbf{BLEU} &  \textbf{Contributed Langs} & \textbf{BLEU}\\
      \hline
      \textbf{ \thead{Individual \\}} & \textcolor{blue-random}{\textbf{gl}} & 13.50 &  \textcolor{blue-random}{\textbf{el}} & 26.82\\
       \hline

       \textbf{\thead{Multi. \\(uniform)\\}} & All langs.  & 22.04 &  All langs.& 26.07 \\\hdashline
      \textbf{ \thead{Multi.\\ (SKD)\\}}  & All langs.& 22.44 &  All langs.& 28.51 \\
      \hline\hline
         \textbf{ \thead{Clus.\\ type1\\}} & \textcolor{blue-random}{\textbf{gl}}, bg, ro, es, it, pt  & 25.53 & \textcolor{blue-random}{\textbf{el}}, ar, he  & 30.05\\\hdashline
      \textbf{\thead{Clus.\\ type2\\}}  &  \thead{\textcolor{blue-random}{\textbf{gl}}, bg, sv, da, nb, de,\\ nl, ro, el, es, it, fr, pt} & 26.81 & \thead{\textcolor{blue-random}{\textbf{el}}, bg, sv, da, nb, de, \\nl, es, it, gl, fr, pt, ro} & 31.35\\\hdashline
      \textbf{\thead{Clus.\\ type3 \\}}& \textcolor{blue-random}{\textbf{gl}}, fr, it, es, pt& 26.93 & \textcolor{blue-random}{\textbf{el}}, mk, bg & 29.13 \\\hdashline
       \textbf{\thead{\\Clus.\\ type4\\}}  & \thead{\textcolor{blue-random}{\textbf{gl}}, fa, ku, eu, hu, et, fi,\\ hy, ka, ar, he, fr, cs, lt,\\ de, nl, it, sv, da, nb, ru,\\ be, pl, bg, sl, pt sr, uk,\\ hr, mk, sk, ro, sq, el, es}& 25.90& \thead{\textcolor{blue-random}{\textbf{el}}, fa, ku, eu, hu, et, fi,\\ hy, ka, ar, he, fr, cs, lt,\\ de, nl, it, sv, da, nb, ru,\\ be, pl, bg, sl, pt, sr, uk,\\ hr, mk, sk, ro, sq, es, gl} & 29.66\\\hdashline
       \thead{\textbf{Avg.}\\}  & -& \textcolor{green}{ \textbf{26.29}} & -& \textcolor{green}{ \textbf{30.04}}\\
       \hline \hline
     \textbf{ \thead{Clus.\\ Rand.1\\} } & \textcolor{blue-random}{\textbf{gl}}, nb, uk, hr, se, ja & 16.53 & \textcolor{blue-random}{\textbf{el}}, id, be & 28.01\\\hdashline
     \textbf{ \thead{Clus.\\ Rand.2\\}} &  \thead{\textcolor{blue-random}{\textbf{gl}}, ta, mk, be, id, sq, pt, \\fr, ur, az, ku, bs, fa } & 20.27 & \thead{\textcolor{blue-random}{\textbf{el}}, cs, lt, id, sk, th, it,\\ hy, ms, hu, mk, my, bn } & 27.78\\\hdashline
     \textbf{ \thead{Clus.\\ Rand.3\\}} & \textcolor{blue-random}{\textbf{gl}}, az, ja, nb, kk & 13.61& \textcolor{blue-random}{\textbf{el}}, sq, th& 27.97\\\hdashline
     \textbf{ \thead{Clus.\\ Rand.4\\}} & \thead{\textcolor{blue-random}{\textbf{gl}}, zh, pt, fa, ar, kk, sr,\\ bg, nl, cs, th, ko, vi, hu,\\ mk, fi, ru, mn, de, sl, el,\\ ka, pl, et, ta, fr, ur, ro,\\ sv, mr, be, bs, uk, sq, az}  &22.20 & \thead{\textcolor{blue-random}{\textbf{el}}, zh, pt, fa, ar, kk, sr,\\ bg, nl, cs,th, ko, vi, hu,\\ mk, fi, ru, mn, de, sl, gl,\\ ka, pl, et, ta, fr, ur, ro,\\ sv, mr, be, bs, uk, sq, az} & 28.82\\\hdashline
      \thead{ \textbf{Avg.}\\}  & - & \textcolor{green}{${\textbf{18.15}}_{\Delta -8.14}$}& - &  \textcolor{green}{${\textbf{28.14}}_{\Delta -1.9}$ }\\
   \hline
    \end{tabular}}
    \caption{Ablation study on using random clusters. Comparison of the (gl$\rightarrow$en) and (el$\rightarrow$en) translation tasks between individual, massive multilingual, and clustering-based multilingual (for actual and random clusters) baselines.}
    \label{tab:random-app}
\end{table*}

\subsection*{Analysis} \label{sec:ap-ana}
\paragraph{Random  clustering  vs  Actual  clustering.} We studied the behaviour of random clusters compared to the actuall clusters in Section 4.1.1. For more clarification about the mentioned discussion, you can refer to Table \ref{tab:random-app}. 
\paragraph{Language Family.} A group of languages that originated from a similar ancestor is known as a \textit{language family}; and a language that does not have any relationship with another languages is called a \textit{language isolate}.. We already discussed the ablation study's results regarding the language families in comparison with isolated languages in Section 4.1.1 of the paper. For better visualisation of results based on language family, we sorted all language pairs based on their language family relations in Tables \ref{tab:clustering-1}, \ref{tab:clustering-2}, \ref{tab:clustering-3}, and \ref{tab:clustering-4}.
\definecolor{c11}{rgb}{0.6, 0.4, 0.8} 
\definecolor{c12}{rgb}{0.0, 0.5, 0.0} 
\definecolor{c13}{rgb}{0.0, 0.0, 1.0} 
\definecolor{c14}{rgb}{0.43, 0.21, 0.1} 
\definecolor{c15}{rgb}{0.13, 0.67, 0.8} 
\definecolor{c16}{rgb}{1.0, 0.44, 0.37} 
\definecolor{c17}{rgb}{0.19, 0.55, 0.91} 
\definecolor{c18}{rgb}{0.54, 0.17, 0.89} 
\definecolor{c19}{rgb}{0.87, 0.36, 0.51} 
\definecolor{c110}{rgb}{1.0, 0.0, 0.5} 
\definecolor{c21}{rgb}{0.0, 0.5, 0.0}
\definecolor{c22}{rgb}{0.0, 0.26, 0.15} 
\definecolor{c23}{rgb}{0.65, 0.16, 0.16} 
\definecolor{c24}{rgb}{1.0, 0.33, 0.64} 
\definecolor{c25}{rgb}{0.28, 0.02, 0.03} 
\definecolor{c26}{rgb}{0.8, 0.33, 0.0} 
\definecolor{c27}{rgb}{0.74, 0.2, 0.64} 
\definecolor{c28}{rgb}{0.5, 0.0, 0.13} 
\definecolor{c29}{rgb}{0.0, 0.42, 0.24} 
\definecolor{c210}{rgb}{0.44, 0.16, 0.39} 
\definecolor{c31}{rgb}{0.89, 0.0, 0.13} 
\definecolor{c32}{rgb}{0.12, 0.3, 0.17} 
\definecolor{c32}{rgb}{0.01, 0.75, 0.24}
\definecolor{c33}{rgb}{0.75, 0.0, 1.0} 
\definecolor{c34}{rgb}{0.89, 0.44, 0.48} 
\definecolor{c35}{rgb}{0.16, 0.32, 0.75} 
\definecolor{c36}{rgb}{0.7, 0.11, 0.11}
\definecolor{c37}{rgb}{0.48, 0.25, 0.0} 
\definecolor{c38}{rgb}{0.0, 0.28, 0.67} 
\definecolor{c39}{rgb}{1.0, 0.22, 0.0}
\definecolor{c310}{rgb}{0.0, 0.18, 0.39}
\definecolor{c311}{rgb}{0.86, 0.08, 0.24}
\definecolor{c41}{rgb}{0.0, 0.0, 0.55}
\definecolor{c42}{rgb}{0.64, 0.0, 0.0}
\definecolor{c43}{rgb}{0.05, 0.5, 0.06}
\newcommand{\mcrot}[4]{\multicolumn{#1}{#2}{\rlap{\rotatebox{#3}{#4}~}}} 
\newcommand*{\twoelementtable}[3][l]%
{%
    \renewcommand{\arraystretch}{0.8}%
    \begin{tabular}[t]{@{}#1@{}}%
        #2\tabularnewline
        #3%
    \end{tabular}%
}

\begin{sidewaystable*}[!htbp] 
    \small
    \hspace{1pt}
\scalebox{0.8}{
\rotatebox{90}{
}
\begin{tabular}
{ ll | cccccccccc|| cccccccccc || ccccccccccc || ccc}\\
    \mcrot{1}{l}{90}{family} & \thead{ lang\\} &   
    \mcrot{1}{l}{90}{\textcolor{c11}{ja, ko, mn, my}} & 
    \mcrot{1}{l}{90}{\textcolor{c12}{ms, th, vi, zh, id}} & 
    \mcrot{1}{l}{90}{\textcolor{c13}{mr, ta, bn, ka}} &
    \mcrot{1}{l}{90}{\textcolor{c14}{ku, fa, kk, eu, hi, ur}} & 
    \mcrot{1}{l}{90}{\textcolor{c15}{el, ar, he}} & 
    \mcrot{1}{l}{90}{\textcolor{c16}{tr, az, fi, hu, hy}} & 
    \mcrot{1}{l}{90}{\textcolor{c17}{uk, pl, ru, mk, lt, be, sk}} & 
    \mcrot{1}{l}{90}{\textcolor{c18}{cs, et, sq, hr, bs, sl, sr}} & 
    \mcrot{1}{l}{90}{\textcolor{c19}{bg, ro, es, gl, it, pt}} &
    \mcrot{1}{l}{90}{\textcolor{c110}{fr, da, sv, nl, de, nb}} & 
    \mcrot{1}{l}{90}{\textcolor{c21}{ko, bn, mr, hi, ur}} & 
    \mcrot{1}{l}{90}{\textcolor{c22}{eu, ar, he}} & 
    \mcrot{1}{l}{90}{\textcolor{c23}{hy, fa, ku}}&
    \mcrot{1}{l}{90}{\textcolor{c24}{hu, tr, az, ja, mn}}&
    \mcrot{1}{l}{90}{\textcolor{c25}{ka, ta}}&
    \mcrot{1}{l}{90}{\textcolor{c26}{kk, my}}&
    \mcrot{1}{l}{90}{\textcolor{c27}{mk, sq, pl, sk, hr, bs, be, et}}&
    \mcrot{1}{l}{90}{\textcolor{c28}{ru, uk, sl, fi, cs, lt}}&
    \mcrot{1}{l}{90}{\textcolor{c29}{zh, th, id, vi, ms}}&
    \mcrot{1}{l}{90}{\textcolor{c210}{bg, sv, da, nb, de, nl, el, es, it, gl, fr, pt, ro}} &
    \mcrot{1}{l}{90}{\textcolor{c31}{et, fi}}&
    \mcrot{1}{l}{90}{\textcolor{c32}{hi, my, hy, ka}}&
    \mcrot{1}{l}{90}{\textcolor{c33}{eu, az, kk, mn, ur, mr, bn, ta, ku, bs, be, ms}}&
    \mcrot{1}{l}{90}{\textcolor{c34}{gl, fr, it, es, pt}}&
    \mcrot{1}{l}{90}{\textcolor{c35}{nb, da, sv}}&
    \mcrot{1}{l}{90}{\textcolor{c36}{de, nl}}&
    \mcrot{1}{l}{90}{\textcolor{c37}{zh, ja, ko, hu, tr}}&
    \mcrot{1}{l}{90}{\textcolor{c38}{lt, sl, hr, sr, cs, sk, pl, ru, uk}}&
    \mcrot{1}{l}{90}{\textcolor{c39}{fa, id, ar, he, th, vi}}&
    \mcrot{1}{l}{90}{\textcolor{c310}{ro, sq}}&
    \mcrot{1}{l}{90}{\textcolor{c311}{mk, bg, el}}&
    \mcrot{1}{l}{90}{\twoelementtable{\textcolor{c41}{fa, ku, eu, hu, et, fi, hy, ka, ar, he, fr, cs, lt, de, nl, sv, da, nb,}}{\textcolor{c41}{ru, be, pl, bg, sl, sr, uk, hr, mk, sk, ro, sq, el, es, gl, pt, it}}} &
    \mcrot{1}{l}{90}{\textcolor{c42}{vi, id, ms, th}}&
     \mcrot{1}{l}{90}{\textcolor{c43}{\thead{kk, az, tr, lt, ur, bn, mr, zh, my, ko, ja, mn, ta\\}}} \\
    \hline 
    & & \multicolumn{10}{c||}{\thead{Clustering Type 1\\}} & \multicolumn{10}{c||}{Clustering Type 2} & \multicolumn{11}{c||}{Clustering Type 3} & 
    \multicolumn{3}{c}{Clustering Type 4}\\
    \hline
    &  & \textcolor{c11}{$c_{1}$} &  \textcolor{c12}{$c_{2}$} &\textcolor{c13}{$c_{3}$} & \textcolor{c14}{$c_{4}$} &\textcolor{c15}{$c_{5}$} & \textcolor{c16}{$c_{6}$} & \textcolor{c17}{$c_{7}$}& \textcolor{c18}{$c_{8}$}&\textcolor{c19}{$c_{9}$}& \textcolor{c110}{$c_{10}$}&
    \textcolor{c21}{\thead{$c_{1}$\\}} &  \textcolor{c22}{$c_{2}$} &\textcolor{c23}{$c_{3}$} & \textcolor{c24}{$c_{4}$} &\textcolor{c25}{$c_{5}$} & \textcolor{c26}{$c_{6}$} & \textcolor{c27}{$c_{7}$}& \textcolor{c28}{$c_{8}$}&\textcolor{c29}{$c_{9}$}& \textcolor{c210}{$c_{10}$}&
    \textcolor{c31}{$c_{1}$} &  \textcolor{c32}{$c_{2}$} &\textcolor{c33}{$c_{3}$} & \textcolor{c34}{$c_{4}$} &\textcolor{c35}{$c_{5}$} & \textcolor{c36}{$c_{6}$} & \textcolor{c37}{$c_{7}$}& \textcolor{c38}{$c_{8}$}&\textcolor{c39}{$c_{9}$}& \textcolor{c310}{$c_{10}$}& \textcolor{c311}{$c_{11}$}&
    \textcolor{c41}{$c_{1}$} &  \textcolor{c42}{$c_{2}$} &\textcolor{c43}{$c_{3}$}\\
    \midrule
    \multirow{13}{*}{\rotatebox{90}{\textbf{IE/Balto-Slavic}}}
    & be  & - & - & - & - & - & -  & \textcolor{c17}{12.7} & - & - & - &
    - & - & - & - & - & - & \textcolor{c27}{ 10.8} & - & - & - & 
    - & - & \textcolor{c33}{8.4} & - & - & - & - & - & -  & - & - & 
    \textcolor{c41}{\textbf{\underline{15.1}} }& -  & - \\
    & bs  & - & - & - & - & - & -  & - &\textcolor{c18}{18.0} & - & - &
    - & - & - & - & - & - & \textcolor{c27}{ 16.8} & - & - & - & 
    - & - & \textcolor{c33}{9.0} & - & - & - & - & - & -  & - & - & 
    \textcolor{c41}{ \textbf{\underline{19.0}}} & -  & - \\
    & sl  & - & - & - & - & - & -  & - & \textcolor{c18}{ 16.4} & - & - &
    - & - & - & - & - & - & - &\textcolor{c28}{ 16.5} & - & - & 
    - & - & - & - & - & - & - &\textcolor{c38}{ 17.7} & -  & - & - & 
   \textcolor{c41}{ \textbf{\underline{18.3}}} & -  & - \\
    & mk  & - & - & - & - & - & -  & \textcolor{c17}{ 21.4} & - & - & - &
    - & - & - & - & - & - &\textcolor{c27}{ 20.0} & - & - & - & 
    - & - & - & - & - & - & - & - & -  & - & \textcolor{c311}{ \textbf{\underline{24.6}}} & 
    \textcolor{c41}{23.8} & -  & - \\
    & lt   & - & - & - & - & - & -  &\textcolor{c17}{ 16.9} & - & - & - &
    - & - & - & - & - & - & - &\textcolor{c28}{16.9} & - & - & 
    - & - & - & - & - & - & - &\textcolor{c38}{ 17.9} & -  & - & - & 
    - & -  &\textcolor{c43}{ \textbf{\underline{18.2}}} \\
    & sk  & - & - & - & - & - & -  & \textcolor{c17}{22.4} & - & - & - &
    - & - & - & - & - & - & - & \textcolor{c28}{ 21.1} & - & - & 
    - & - & - & - & - & - & - & \textcolor{c38}{ \textbf{\underline{24.0}}} & -  & - & - & 
    \textcolor{c41}{ 23.7} & -  & - \\
    & cs  & - & - & - & - & - & -  & - &\textcolor{c18}{ 21.7} & - & - &
    - & - & - & - & - & - & - & \textcolor{c28}{ 22.4} & - & - & 
    - & - & - & - & - & - & - & \textcolor{c38}{ \textbf{\underline{23.1}}}& -  & - & - & 
    \textcolor{c41}{ 22.8} & -  & - \\
    & uk  & - & - & - & - & - & -  & \textcolor{c17}{ 23.0 }& - & - & - &
    - & - & - & - & - & - & - &  \textcolor{c28}{ 22.9 }& - & - & 
    - & - & - & - & - & - & - & \textcolor{c38}{23.5} & -  & - & - & 
     \textcolor{c41}{\textbf{\underline{23.6}}} & -  & - \\
    & hr  & - & - & - & - & - & -  & - & \textcolor{c18}{ 27.5} & - & - &
    - & - & - & - & - & - &  \textcolor{c27}{25.6} & - & - & - & 
    - & - & - & - & - & - & - & \textcolor{c38}{ 28.6} & -  & - & - & 
    \textcolor{c41}{28.3} & -  & - \\
    & sr   & - & - & - & - & - & -  & - & \textcolor{c18}{ 27.4} & - & - &
    - & - & - & - & - & - & \textcolor{c27}{25.7} & - & - & - & 
    - & - & - & - & - & - & - & \textcolor{c38}{ \textbf{\underline{27.6}}} & -  & - & - & 
    \textcolor{c41}{27.1} & -  & - \\
    & bg  & - & - & - & - & - & -  & - & - &  \textcolor{c19}{31.6} & - &
    - & - & - & - & - & - & - & - & - &\textcolor{c210}{ \textbf{\underline{32.1}}}& 
    - & - & - & - & - & - & - & - & -  & - & \textcolor{c311}{29.8} & 
    \textcolor{c41}{30.4} & -  & - \\
    & pl  & - & - & - & - & - & -  & \textcolor{c17}{ 19.4} & - & - & - &
    - & - & - & - & - & - & \textcolor{c27}{ 18.3} & - & - & - & 
    - & - & - & - & - & - & - & \textcolor{c38}{ 19.9} & -  & - & - & 
    \textcolor{c41}{\textbf{\underline{20.2}}} & -  & - \\
    & ru  & - & - & - & - & - & -  &  \textcolor{c17}{20.8} & - & - & - &
    - & - & - & - & - & - & - & \textcolor{c28}{20.8} & - & - & 
    - & - & - & - & - & - & - & \textcolor{c38}{ 21.2} & -  & - & - & 
    \textcolor{c41}{\textbf{\underline{21.4}}} & -  & - \\
    \midrule
    \multirow{6}{*}{\rotatebox{90}{\textbf{\smaller {IE/Indo-Iranian}}}}
    & bn  & - & - & \textcolor{c13}{ 9.1} & - & - & -  & - & - & - & - &
    \textcolor{c21}{10.1} & - & - & - & - & - & - & - & - & - & 
    - & - & \textcolor{c33}{\textbf{\underline{12.1}}} & - & - & - & - & - & -  & - & - & 
    - & -  & \textcolor{c43}{10.1} \\
    & ur   & - & - & - & \textcolor{c14}{ 13.3} & - & -  & - & - & - & - &
    \textcolor{c21}{13.1} & - & - & - & - & - & - & - & - & - & 
    - & - & \textcolor{c33}{12.0} & - & - & - & - & - & -  & - & - & 
    - & -  & \textcolor{c43}{\textbf{\underline{13.3}}} \\
    & ku  & - & - & - & \textcolor{c14}{ 9.4} & - & -  & - & - & - & - &
   \textcolor{c21}{ 6.9} & - & - & - & - & - & - & - & - & - & 
    - & - & \textcolor{c33}{ 9.2} & - & - & - & - & - & -  & - & - & 
    - & -  &  \textcolor{c43}{\textbf{\underline{12.9}}} \\
    & hi  & - & - & - & \textcolor{c14}{ \textbf{\underline{13.2}}} & - & -  & - & - & - & - &
    \textcolor{c21}{12.0} & - & - & - & - & - & - & - & - & - & 
    - & \textcolor{c32}{12.1} & - & - & - & - & - & - & -  & - & - & 
   \textcolor{c41}{ 12.8} & -  & - \\
    & fa   & - & - & - & \textcolor{c14}{ 21.3} & - & -  & - & - & - & - &
    - & - & \textcolor{c23}{ 22.4} & - & - & - & - & - & - & - & 
    - & - & - & - & - & - & - & - & \textcolor{c39}{ \textbf{\underline{23.5}}} & - & - & 
    \textcolor{c41}{ 22.2} & -  & - \\
    & mr   & - & - & \textcolor{c13}{ 8.7} & - & - & -  & - & - & - & - &
   \textcolor{c21}{ 9.0} & - & - & - & - & - & - & - & - & - & 
    - & - & \textcolor{c33}{ 8.3} & - & - & - & - & - & - & - & - & 
    - & -  & \textcolor{c43}{ \textbf{\underline{9.0}}} \\
    \hline
    \multirow{6}{*}{\rotatebox{90}{\textbf{\smaller {IE/Italic}}}}
    & gl  & - & - & - & - & - & -  & - & - & \textcolor{c19}{ 25.5} & - &
    - & - & - & - & - & - & - & - & - & \textcolor{c210}{ 26.8} & 
    - & - & - & \textcolor{c34}{ \textbf{\underline{26.9}}}& - & - & - & - & -  & - & - & 
    \textcolor{c41}{25.9} & -  & - \\
    & pt  & - & - & - & - & - & -  & - & - & \textcolor{c19}{ 33.3} & - &
    - & - & - & - & - & - & - & - & - & \textcolor{c210}{ 33.8} & 
    - & - & - & \textcolor{c34}{ 33.2} & - & - & - & - & -  & - & - & 
   \textcolor{c41}{ 32.5} & -  & - \\
    & ro  & - & - & - & - & - & -  & - & - & \textcolor{c19}{ 28.0} & - &
    - & - & - & - & - & - & - & - & - & \textcolor{c210}{ 28.4} & 
    - & - & - & - & - & - & - & - & -  & \textcolor{c310}{ \textbf{\underline{28.9}}} & - & 
    \textcolor{c41}{27.3} & -  & - \\
    & fr  & - & - & - & - & - & -  & - & - & - & \textcolor{c110}{ \textbf{\underline{32.2}}} &
    - & - & - & - & - & - & - & - & - & \textcolor{c210}{ 32.1} & 
    - & - & - & \textcolor{c34}{ 31.1} & - & - & - & - & -  & - & - & 
    \textcolor{c41}{ 31.3} & -  & - \\
    & es  & - & - & - & - & - & -  & - & - & \textcolor{c19}{ 29.0} & - &
    - & - & - & - & - & - & - & - & - & \textcolor{c210}{ \textbf{\underline{33.4}}} & 
    - & - & - & \textcolor{c34}{ 31.4} & - & - & - & - & -  & - & - & 
   \textcolor{c41}{ 31.8} & -  & - \\
    & it  & - & - & - & - & - & -  & - & - & \textcolor{c19}{ 30.5} & - &
    - & - & - & - & - & - & - & - & - & \textcolor{c210}{\textbf{\underline{30.8}}} & 
    - & - & - & \textcolor{c34}{ 30.3} & - & - & - & - & -  & - & - & 
    \textcolor{c41}{ 29.4} & -  & - \\
    \bottomrule
\end{tabular} 
}
\caption{The results of clustering-based multilingual NMT with knowledge distillation for languages belong to IE/Balto-Slavic, IE/Indo-Iranina, and IE/Italic.}
\label{tab:clustering-1}
\end{sidewaystable*}
\begin{sidewaystable*}[!htbp]
    \small
    \hspace{1pt}
\scalebox{0.74}{
\rotatebox{0}{
\begin{tabular}
{ ll | cccccccccc|| cccccccccc || ccccccccccc || ccc}\\
    \mcrot{1}{l}{90}{family} & \thead{ lang\\} &   
    \mcrot{1}{l}{90}{\textcolor{c11}{ja, ko, mn, my}} & 
    \mcrot{1}{l}{90}{\textcolor{c12}{ms, th, vi, zh, id}} & 
    \mcrot{1}{l}{90}{\textcolor{c13}{mr, ta, bn, ka}} &
    \mcrot{1}{l}{90}{\textcolor{c14}{ku, fa, kk, eu, hi, ur}} & 
    \mcrot{1}{l}{90}{\textcolor{c15}{el, ar, he}} & 
    \mcrot{1}{l}{90}{\textcolor{c16}{tr, az, fi, hu, hy}} & 
    \mcrot{1}{l}{90}{\textcolor{c17}{uk, pl, ru, mk, lt, be, sk}} & 
    \mcrot{1}{l}{90}{\textcolor{c18}{cs, et, sq, hr, bs, sl, sr}} & 
    \mcrot{1}{l}{90}{\textcolor{c19}{bg, ro, es, gl, it, pt}} &
    \mcrot{1}{l}{90}{\textcolor{c110}{fr, da, sv, nl, de, nb}} & 
    \mcrot{1}{l}{90}{\textcolor{c21}{ko, bn, mr, hi, ur}} & 
    \mcrot{1}{l}{90}{\textcolor{c22}{eu, ar, he}} & 
    \mcrot{1}{l}{90}{\textcolor{c23}{hy, fa, ku}}&
    \mcrot{1}{l}{90}{\textcolor{c24}{hu, tr, az, ja, mn}}&
    \mcrot{1}{l}{90}{\textcolor{c25}{ka, ta}}&
    \mcrot{1}{l}{90}{\textcolor{c26}{kk, my}}&
    \mcrot{1}{l}{90}{\textcolor{c27}{mk, sq, pl, sk, hr, bs, be, et}}&
    \mcrot{1}{l}{90}{\textcolor{c28}{ru, uk, sl, fi, cs, lt}}&
    \mcrot{1}{l}{90}{\textcolor{c29}{zh, th, id, vi, ms}}&
    \mcrot{1}{l}{90}{\textcolor{c210}{bg, sv, da, nb, de, nl, el, es, it, gl, fr, pt, ro}} &
    \mcrot{1}{l}{90}{\textcolor{c31}{et, fi}}&
    \mcrot{1}{l}{90}{\textcolor{c32}{hi, my, hy, ka}}&
    \mcrot{1}{l}{90}{\textcolor{c33}{eu, az, kk, mn, ur, mr, bn, ta, ku, bs, be, ms}}&
    \mcrot{1}{l}{90}{\textcolor{c34}{gl, fr, it, es, pt}}&
    \mcrot{1}{l}{90}{\textcolor{c35}{nb, da, sv}}&
    \mcrot{1}{l}{90}{\textcolor{c36}{de, nl}}&
    \mcrot{1}{l}{90}{\textcolor{c37}{zh, ja, ko, hu, tr}}&
    \mcrot{1}{l}{90}{\textcolor{c38}{lt, sl, hr, sr, cs, sk, pl, ru, uk}}&
    \mcrot{1}{l}{90}{\textcolor{c39}{fa, id, ar, he, th, vi}}&
    \mcrot{1}{l}{90}{\textcolor{c310}{ro, sq}}&
    \mcrot{1}{l}{90}{\textcolor{c311}{mk, bg, el}}&
    \mcrot{1}{l}{90}{\twoelementtable{\textcolor{c41}{fa, ku, eu, hu, et, fi, hy, ka, ar, he, fr, cs, lt, de, nl, sv, da, nb,}}{\textcolor{c41}{ru, be, pl, bg, sl, sr, uk, hr, mk, sk, ro, sq, el, es, gl, pt, it}}} &
    \mcrot{1}{l}{90}{\textcolor{c42}{vi, id, ms, th}}&
     \mcrot{1}{l}{90}{\textcolor{c43}{\thead{kk, az, tr, lt, ur, bn, mr, zh, my, ko, ja, mn, ta\\}}} \\
    \hline 
    & & \multicolumn{10}{c||}{\thead{Clustering Type 1\\}} & \multicolumn{10}{c||}{Clustering Type 2} & \multicolumn{11}{c||}{Clustering Type 3} & 
    \multicolumn{3}{c}{Clustering Type 4}\\
    \hline
    &  & \textcolor{c11}{$c_{1}$} &  \textcolor{c12}{$c_{2}$} &\textcolor{c13}{$c_{3}$} & \textcolor{c14}{$c_{4}$} &\textcolor{c15}{$c_{5}$} & \textcolor{c16}{$c_{6}$} & \textcolor{c17}{$c_{7}$}& \textcolor{c18}{$c_{8}$}&\textcolor{c19}{$c_{9}$}& \textcolor{c110}{$c_{10}$}&
    \textcolor{c21}{\thead{$c_{1}$\\}} &  \textcolor{c22}{$c_{2}$} &\textcolor{c23}{$c_{3}$} & \textcolor{c24}{$c_{4}$} &\textcolor{c25}{$c_{5}$} & \textcolor{c26}{$c_{6}$} & \textcolor{c27}{$c_{7}$}& \textcolor{c28}{$c_{8}$}&\textcolor{c29}{$c_{9}$}& \textcolor{c210}{$c_{10}$}&
    \textcolor{c31}{$c_{1}$} &  \textcolor{c32}{$c_{2}$} &\textcolor{c33}{$c_{3}$} & \textcolor{c34}{$c_{4}$} &\textcolor{c35}{$c_{5}$} & \textcolor{c36}{$c_{6}$} & \textcolor{c37}{$c_{7}$}& \textcolor{c38}{$c_{8}$}&\textcolor{c39}{$c_{9}$}& \textcolor{c310}{$c_{10}$}& \textcolor{c311}{$c_{11}$}&
    \textcolor{c41}{$c_{1}$} &  \textcolor{c42}{$c_{2}$} &\textcolor{c43}{$c_{3}$}\\
    \midrule
    \multirow{5}{*}{\rotatebox{90}{\textbf{IE/Germanic}}}
    & nb  & - & - & - & - & - & -  & - & - & - & \textcolor{c110}{ 33.5} &
    - & - & - & - & - & - & - & - & - &\textcolor{c210}{\textbf{\underline{34.3}}} & 
    - & - & - & - &\textcolor{c35}{ 28.7} & - & - & - & -  & - & - & 
    \textcolor{c41}{30.8} & -  & - \\
    & da  & - & - & - & - & - & -  & - & - & - &\textcolor{c110}{35.0} &
    - & - & - & - & - & - & - & - & - & \textcolor{c210}{\textbf{\underline{39.7}}} & 
    - & - & - & - & \textcolor{c35}{ 30.5} & - & - & - & -  & - & - & 
    \textcolor{c41}{32.0} & -  & - \\
    & sv  & - & - & - & - & - & -  & - & - & - &\textcolor{c110}{ 31.3} &
    - & - & - & - & - & - & - & - & - &\textcolor{c210}{ \textbf{\underline{34.5}}} & 
    - & - & - & - & \textcolor{c35}{26.8} & - & - & - & -  & - & - & 
   \textcolor{c41}{28.6} & -  & - \\
    & de  & - & - & - & - & - & -  & - & - & - & \textcolor{c110}{16.8} &
    - & - & - & - & - & - & - & - & - & \textcolor{c210}{\textbf{\underline{17.7}}} & 
    - & - & - & - & - &\textcolor{c36}{15.4} & - & - & -  & - & - & 
    \textcolor{c41}{16.6} & -  & - \\
    & nl  & - & - & - & - & - & -  & - & - & - & \textcolor{c110}{28.0} &
    - & - & - & - & - & - & - & - & - & \textcolor{c210}{29.0} & 
    - & - & - & - & - & \textcolor{c36}{\textbf{\underline{29.1}}} & - & - & -  & - & - & 
    \textcolor{c41}{28.0} & -  & - \\
    \midrule
    \multirow{3}{*}{\rotatebox{90}{\textbf{Turkic}}}
    & kk  & - & - & - & \textcolor{c14}{ 7.0} & - & -  & - & - & - & - &
    - & - & - & - & - & \textcolor{c26}{3.5} & - & - & - & - & 
    - & - &  \textcolor{c33}{3.1} & - & - & - & - & - & -  & - & - & 
    - & -  &  \textcolor{c43}{\textbf{\underline{8.1}}} \\
    & az  & - & - & - & - & - & \textcolor{c16}{\textbf{\underline{9.5}}}  & - & - & - & - &
    - & - & - & \textcolor{c24}{9.1} & - & - & - & - & - & - & 
    - & - & \textcolor{c33}{8.6} & - & - & - & - & - & -  & - & - & 
    - & -  &\textcolor{c43}{9.2} \\
    & tr  & - & - & - & - & - & \textcolor{c16}{18.6}  & - & - & - & - &
    - & - & - & \textcolor{c24}{18.1} & - & - & - & - & - & - & 
    - & - & - & - & - & - & \textcolor{c37}{\textbf{19.8}} & - & -  & - & - & 
    - & -  &  \textcolor{c43}{18.2} \\
     \midrule
    \multirow{3}{*}{\rotatebox{90}{\textbf{Uralic}}}
    & et  & - & - & - & - & - & -  & - & \textcolor{c18}{13.4} & - & - &
    - & - & - & - & - & - & \textcolor{c27}{13.2} & - & - & - & 
    \textcolor{c31}{10.4} & - & - & - & - & - & - & - & -  & - & - & 
    \textcolor{c41}{13.9} & -  & - \\
    & fi  & - & - & - & - & - & \textcolor{c16}{ 11.3}  & - & - & - & - &
    - & - & - & - & - & - & - & \textcolor{c28}{12.5} & - & - & 
    \textcolor{c31}{10.5} & - & - & - & - & - & - & - & -  & - & - & 
   \textcolor{c41}{\textbf{\underline{12.7}}} & -  & - \\
    & hu  & - & - & - & - & - & \textcolor{c16}{19.0}  & - & - & - & - &
    - & - & - & \textcolor{c24}{18.4} & - & - & - & - & - & - & 
    - & - & - & - & - & - &  \textcolor{c37}{\textbf{\underline{20.1}}} & - & -  & - & - & 
    \textcolor{c41}{19.8} & -  & - \\
     \midrule
    \multirow{4}{*}{\rotatebox{90}{ \textbf{Afroasiatic}}}
     & & & & & & & & & & & & & & & & & & & & & & & & & & & & & & & & & & & \\
    & he  & - & - & - & - &  \textcolor{c15}{ \textbf{\underline{32.8}}}  & -  & - & - & - & - &
    - & - & - & - & - & - & \textcolor{c27}{32.1} & - & - & - & 
   - &  \textcolor{c32}{31.3} & - & - & - & - & - & - & -  & - & - & 
    \textcolor{c41}{30.0} & -  & - \\ 
    & ar  & - & - & - & - & \textcolor{c15}{\textbf{\underline{28.5}}} & - & - & - & - & - &
    - & \textcolor{c22}{28.0} & - & - & - & - & - & - & - & - & 
    - & - & - & - & - & - & - & - & \textcolor{c39}{27.4}  & - & - & 
   \textcolor{c41}{25.8} & -  & - \\
     & & & & & & & & & & & & & & & & & & & & & & & & & & & & & & & & & & & \\
     \midrule
    \multirow{5}{*}{\rotatebox{90}{\small {\textbf{Sino-Tibetan}}}}
     & & & & & & & & & & & & & & & & & & & & & & & & & & & & & & & & & & & \\
    & my  & \textcolor{c11}{9.6} & - & - & - & - & -  & - & - & - & - &
    - & - & - & - & - & \textcolor{c26}{ \textbf{\underline{6.3}}} & -  & - & - & - & 
   - &  \textcolor{c32}{8.4} & - & - & - & - & - & - & -  & - & - & 
    \textcolor{c41}{9.5} & -  & - \\
     & & & & & & & & & & & & & & & & & & & & & & & & & & & & & & & & & & & \\
    & zh  & - & \textcolor{c12}{22.1} & - & - & - & -  & - & - & - & - &
    - & - & - & - & - & - & - & - & \textcolor{c29}{22.1} & - & 
  - & - & - & - & - & - &  \textcolor{c37}{23.9} & - & -  & - & - & 
   - & -  &  \textcolor{c43}{22.8} \\
     & & & & & & & & & & & & & & & & & & & & & & & & & & & & & & & & & & & \\
  
     \midrule
    \multirow{5}{*}{\rotatebox{90}{\small{\textbf{Austronesian}}}}
     & & & & & & & & & & & & & & & & & & & & & & & & & & & & & & & & & & & \\
    & ms  & - & \textcolor{c12}{12.9} & - & - & - & -  & - &- & - & - &
    - & - & - & - & - & - &- & - &  \textcolor{c29}{12.9} & - & 
    - & - & \textcolor{c33}{7.6} & - & - & - & - & - & -  & - & - & 
   - &  \textcolor{c42}{\textbf{\underline{12.9}}}  & - \\ 
     & & & & & & & & & & & & & & & & & & & & & & & & & & & & & & & & & & & \\
    & id  & - & \textcolor{c12}{21.1} & - & - & - & -  & - & - & - & - &
    - & - & - & - & - & - & - & - & \textcolor{c29}{21.1} & - & 
   - & - & - & - & - & - & - & - & \textcolor{c39}{22.5}  & - & - & 
    - & \textcolor{c42}{21.1}  & - \\
     & & & & & & & & & & & & & & & & & & & & & & & & & & & & & & & & & & & \\
    \bottomrule
\end{tabular} 
}
}
\caption{The results of clustering-based multilingual NMT with knowledge distillation for languages belong to IE/Germanic, Turkic, Uralic, Afroasiatic, Sino-Tibetan, and Austronesian.}
\label{tab:clustering-2}
\end{sidewaystable*}
\begin{sidewaystable*}[!htbp] 
    \small
    \hspace{1pt}
\scalebox{0.78}{
\rotatebox{0}{
\begin{tabular}
{ ll | cccccccccc|| cccccccccc || ccccccccccc || ccc}\\
    \mcrot{1}{l}{90}{family} & \thead{ lang\\} &   
    \mcrot{1}{l}{90}{\textcolor{c11}{ja, ko, mn, my}} & 
    \mcrot{1}{l}{90}{\textcolor{c12}{ms, th, vi, zh, id}} & 
    \mcrot{1}{l}{90}{\textcolor{c13}{mr, ta, bn, ka}} &
    \mcrot{1}{l}{90}{\textcolor{c14}{ku, fa, kk, eu, hi, ur}} & 
    \mcrot{1}{l}{90}{\textcolor{c15}{el, ar, he}} & 
    \mcrot{1}{l}{90}{\textcolor{c16}{tr, az, fi, hu, hy}} & 
    \mcrot{1}{l}{90}{\textcolor{c17}{uk, pl, ru, mk, lt, be, sk}} & 
    \mcrot{1}{l}{90}{\textcolor{c18}{cs, et, sq, hr, bs, sl, sr}} & 
    \mcrot{1}{l}{90}{\textcolor{c19}{bg, ro, es, gl, it, pt}} &
    \mcrot{1}{l}{90}{\textcolor{c110}{fr, da, sv, nl, de, nb}} & 
    \mcrot{1}{l}{90}{\textcolor{c21}{ko, bn, mr, hi, ur}} & 
    \mcrot{1}{l}{90}{\textcolor{c22}{eu, ar, he}} & 
    \mcrot{1}{l}{90}{\textcolor{c23}{hy, fa, ku}}&
    \mcrot{1}{l}{90}{\textcolor{c24}{hu, tr, az, ja, mn}}&
    \mcrot{1}{l}{90}{\textcolor{c25}{ka, ta}}&
    \mcrot{1}{l}{90}{\textcolor{c26}{kk, my}}&
    \mcrot{1}{l}{90}{\textcolor{c27}{mk, sq, pl, sk, hr, bs, be, et}}&
    \mcrot{1}{l}{90}{\textcolor{c28}{ru, uk, sl, fi, cs, lt}}&
    \mcrot{1}{l}{90}{\textcolor{c29}{zh, th, id, vi, ms}}&
    \mcrot{1}{l}{90}{\textcolor{c210}{bg, sv, da, nb, de, nl, el, es, it, gl, fr, pt, ro}} &
    \mcrot{1}{l}{90}{\textcolor{c31}{et, fi}}&
    \mcrot{1}{l}{90}{\textcolor{c32}{hi, my, hy, ka}}&
    \mcrot{1}{l}{90}{\textcolor{c33}{eu, az, kk, mn, ur, mr, bn, ta, ku, bs, be, ms}}&
    \mcrot{1}{l}{90}{\textcolor{c34}{gl, fr, it, es, pt}}&
    \mcrot{1}{l}{90}{\textcolor{c35}{nb, da, sv}}&
    \mcrot{1}{l}{90}{\textcolor{c36}{de, nl}}&
    \mcrot{1}{l}{90}{\textcolor{c37}{zh, ja, ko, hu, tr}}&
    \mcrot{1}{l}{90}{\textcolor{c38}{lt, sl, hr, sr, cs, sk, pl, ru, uk}}&
    \mcrot{1}{l}{90}{\textcolor{c39}{fa, id, ar, he, th, vi}}&
    \mcrot{1}{l}{90}{\textcolor{c310}{ro, sq}}&
    \mcrot{1}{l}{90}{\textcolor{c311}{mk, bg, el}}&
    \mcrot{1}{l}{90}{\twoelementtable{\textcolor{c41}{fa, ku, eu, hu, et, fi, hy, ka, ar, he, fr, cs, lt, de, nl, sv, da, nb,}}{\textcolor{c41}{ru, be, pl, bg, sl, sr, uk, hr, mk, sk, ro, sq, el, es, gl, pt, it}}} &
    \mcrot{1}{l}{90}{\textcolor{c42}{vi, id, ms, th}}&
     \mcrot{1}{l}{90}{\textcolor{c43}{\thead{kk, az, tr, lt, ur, bn, mr, zh, my, ko, ja, mn, ta\\}}} \\
    \hline 
    & & \multicolumn{10}{c||}{\thead{Clustering Type 1\\}} & \multicolumn{10}{c||}{Clustering Type 2} & \multicolumn{11}{c||}{Clustering Type 3} & 
    \multicolumn{3}{c}{Clustering Type 4}\\
    \hline
    &  & \textcolor{c11}{$c_{1}$} &  \textcolor{c12}{$c_{2}$} &\textcolor{c13}{$c_{3}$} & \textcolor{c14}{$c_{4}$} &\textcolor{c15}{$c_{5}$} & \textcolor{c16}{$c_{6}$} & \textcolor{c17}{$c_{7}$}& \textcolor{c18}{$c_{8}$}&\textcolor{c19}{$c_{9}$}& \textcolor{c110}{$c_{10}$}&
    \textcolor{c21}{\thead{$c_{1}$\\}} &  \textcolor{c22}{$c_{2}$} &\textcolor{c23}{$c_{3}$} & \textcolor{c24}{$c_{4}$} &\textcolor{c25}{$c_{5}$} & \textcolor{c26}{$c_{6}$} & \textcolor{c27}{$c_{7}$}& \textcolor{c28}{$c_{8}$}&\textcolor{c29}{$c_{9}$}& \textcolor{c210}{$c_{10}$}&
    \textcolor{c31}{$c_{1}$} &  \textcolor{c32}{$c_{2}$} &\textcolor{c33}{$c_{3}$} & \textcolor{c34}{$c_{4}$} &\textcolor{c35}{$c_{5}$} & \textcolor{c36}{$c_{6}$} & \textcolor{c37}{$c_{7}$}& \textcolor{c38}{$c_{8}$}&\textcolor{c39}{$c_{9}$}& \textcolor{c310}{$c_{10}$}& \textcolor{c311}{$c_{11}$}&
    \textcolor{c41}{$c_{1}$} &  \textcolor{c42}{$c_{2}$} &\textcolor{c43}{$c_{3}$}\\
    \midrule
    \multirow{3}{*}{\rotatebox{90}{\small{\textbf{Koreanic}}}}
     & & & & & & & & & & & & & & & & & & & & & & & & & & & & & & & & & & & \\
    & ko  & \textcolor{c11}{15.1} & - & - & - & - & -  & - & - & - & - &
    \textcolor{c21}{15.1} & - & - & - & - & - & - & - & - & - & 
   - & - & - & - & - & - & \textcolor{c37}{\textbf{\underline{17.4}}} & - & -  & - & - & 
   - & -  & \textcolor{c43}{16.0} \\ 
     & & & & & & & & & & & & & & & & & & & & & & & & & & & & & & & & & & & \\
     \midrule
    \multirow{3}{*}{\rotatebox{90}{\small{\textbf{Japonic}}}}
     & & & & & & & & & & & & & & & & & & & & & & & & & & & & & & & & & & & \\
    & ja  & \textcolor{c11}{10.1} & - & - & - & - & -  & - & - & - & - &
    - & - & - & \textcolor{c24}{10.1} & - & - &- & - & - & - & 
    - & - & - & - & - & - & \textcolor{c37}{10.1} & - & -  & - & - & 
   - & -  & \textcolor{c43}{8.6} \\ 
     & & & & & & & & & & & & & & & & & & & & & & & & & & & & & & & & & & & \\

     \midrule
    \multirow{5}{*}{\rotatebox{90}{\small{\textbf{Austroasiatic}}}}
     & & & & & & & & & & & & & & & & & & & & & & & & & & & & & & & & & & & \\
      & & & & & & & & & & & & & & & & & & & & & & & & & & & & & & & & & & & \\
    & vi  & - & \textcolor{c12}{21.6} & - & - & - & -  & - &- & - & - &
    - & - & - & - & - & - &  - & - & \textcolor{c29}{21.6} & - & 
   - & - & - & - & - & - & - & - & \textcolor{c39}{\textbf{\underline{21.3}}}  & - & - & 
    - & \textcolor{c42}{20.3}  & - \\ 
     & & & & & & & & & & & & & & & & & & & & & & & & & & & & & & & & & & & \\
      & & & & & & & & & & & & & & & & & & & & & & & & & & & & & & & & & & & \\
     \midrule
    \multirow{5}{*}{\rotatebox{90}{\small{\textbf{IE/Hellenic}}}}
     & & & & & & & & & & & & & & & & & & & & & & & & & & & & & & & & & & & \\
      & & & & & & & & & & & & & & & & & & & & & & & & & & & & & & & & & & & \\
    & el  & - & - & - & - & \textcolor{c15}{30.0} & -  & - & - & - & - &
    - & - & - & - & - & - & - & - & - & \textcolor{c210}{\textbf{\underline{31.3}}} & 
    - & - & - & - & - & - & - & - & -  & - & \textcolor{c311}{29.1} & 
   \textcolor{c41}{29.6} & -  & - \\ 
     & & & & & & & & & & & & & & & & & & & & & & & & & & & & & & & & & & & \\
      & & & & & & & & & & & & & & & & & & & & & & & & & & & & & & & & & & & \\
     \midrule
    \multirow{3}{*}{\rotatebox{90}{\small{\textbf{Kra-Dai }}}}
     & & & & & & & & & & & & & & & & & & & & & & & & & & & & & & & & & & & \\
    & th  & - & \textcolor{c12}{22.9} & - & - & - & -  & - & - & - & - &
    - & - & - & - & - & - & - & - & \textcolor{c29}{\textbf{\underline{22.9}}} & - & 
   - & - & - & - & - & - & - & - & \textcolor{c39}{23.09} & - & - & 
    - & \textcolor{c42}{22.8} & - \\ 
     & & & & & & & & & & & & & & & & & & & & & & & & & & & & & & & & & & & \\
     \midrule
    \multirow{5}{*}{\rotatebox{90}{\small{\textbf{IE/Albanian}}}}
     & & & & & & & & & & & & & & & & & & & & & & & & & & & & & & & & & & & \\
      & & & & & & & & & & & & & & & & & & & & & & & & & & & & & & & & & & & \\
    & sq  & - & - & - & - & - & -  & - &\textcolor{c18}{24.7} & - & - &
    - & - & - & - & - & - &\textcolor{c27}{23.2} & - & - & - & 
    - & - & - & - & - & - & - & - & -  & \textcolor{c310}{\textbf{\underline{26.8}}} & - & 
    \textcolor{c41}{26.4} & -  & - \\ 
     & & & & & & & & & & & & & & & & & & & & & & & & & & & & & & & & & & & \\
      & & & & & & & & & & & & & & & & & & & & & & & & & & & & & & & & & & & \\
    \bottomrule
\end{tabular} 
}}
\caption{The results of clustering-based multilingual NMT with knowledge distillation for languages belong to Koreanic, Japonic, Austroasiatic, IE/Hellenic, Kra-Dai, and IE/Albanian families.}
\label{tab:clustering-3}
\end{sidewaystable*}
\begin{sidewaystable*}[!htbp]
    \small
    \hspace{1pt}
\scalebox{0.81}{
\rotatebox{0}{
\begin{tabular}
{ ll | cccccccccc|| cccccccccc || ccccccccccc || ccc}\\
    \mcrot{1}{l}{90}{family} & \thead{ lang\\} &   
    \mcrot{1}{l}{90}{\textcolor{c11}{ja, ko, mn, my}} & 
    \mcrot{1}{l}{90}{\textcolor{c12}{ms, th, vi, zh, id}} & 
    \mcrot{1}{l}{90}{\textcolor{c13}{mr, ta, bn, ka}} &
    \mcrot{1}{l}{90}{\textcolor{c14}{ku, fa, kk, eu, hi, ur}} & 
    \mcrot{1}{l}{90}{\textcolor{c15}{el, ar, he}} & 
    \mcrot{1}{l}{90}{\textcolor{c16}{tr, az, fi, hu, hy}} & 
    \mcrot{1}{l}{90}{\textcolor{c17}{uk, pl, ru, mk, lt, be, sk}} & 
    \mcrot{1}{l}{90}{\textcolor{c18}{cs, et, sq, hr, bs, sl, sr}} & 
    \mcrot{1}{l}{90}{\textcolor{c19}{bg, ro, es, gl, it, pt}} &
    \mcrot{1}{l}{90}{\textcolor{c110}{fr, da, sv, nl, de, nb}} & 
    \mcrot{1}{l}{90}{\textcolor{c21}{ko, bn, mr, hi, ur}} & 
    \mcrot{1}{l}{90}{\textcolor{c22}{eu, ar, he}} & 
    \mcrot{1}{l}{90}{\textcolor{c23}{hy, fa, ku}}&
    \mcrot{1}{l}{90}{\textcolor{c24}{hu, tr, az, ja, mn}}&
    \mcrot{1}{l}{90}{\textcolor{c25}{ka, ta}}&
    \mcrot{1}{l}{90}{\textcolor{c26}{kk, my}}&
    \mcrot{1}{l}{90}{\textcolor{c27}{mk, sq, pl, sk, hr, bs, be, et}}&
    \mcrot{1}{l}{90}{\textcolor{c28}{ru, uk, sl, fi, cs, lt}}&
    \mcrot{1}{l}{90}{\textcolor{c29}{zh, th, id, vi, ms}}&
    \mcrot{1}{l}{90}{\textcolor{c210}{bg, sv, da, nb, de, nl, el, es, it, gl, fr, pt, ro}} &
    \mcrot{1}{l}{90}{\textcolor{c31}{et, fi}}&
    \mcrot{1}{l}{90}{\textcolor{c32}{hi, my, hy, ka}}&
    \mcrot{1}{l}{90}{\textcolor{c33}{eu, az, kk, mn, ur, mr, bn, ta, ku, bs, be, ms}}&
    \mcrot{1}{l}{90}{\textcolor{c34}{gl, fr, it, es, pt}}&
    \mcrot{1}{l}{90}{\textcolor{c35}{nb, da, sv}}&
    \mcrot{1}{l}{90}{\textcolor{c36}{de, nl}}&
    \mcrot{1}{l}{90}{\textcolor{c37}{zh, ja, ko, hu, tr}}&
    \mcrot{1}{l}{90}{\textcolor{c38}{lt, sl, hr, sr, cs, sk, pl, ru, uk}}&
    \mcrot{1}{l}{90}{\textcolor{c39}{fa, id, ar, he, th, vi}}&
    \mcrot{1}{l}{90}{\textcolor{c310}{ro, sq}}&
    \mcrot{1}{l}{90}{\textcolor{c311}{mk, bg, el}}&
    \mcrot{1}{l}{90}{\twoelementtable{\textcolor{c41}{fa, ku, eu, hu, et, fi, hy, ka, ar, he, fr, cs, lt, de, nl, sv, da, nb,}}{\textcolor{c41}{ru, be, pl, bg, sl, sr, uk, hr, mk, sk, ro, sq, el, es, gl, pt, it}}} &
    \mcrot{1}{l}{90}{\textcolor{c42}{vi, id, ms, th}}&
     \mcrot{1}{l}{90}{\textcolor{c43}{\thead{kk, az, tr, lt, ur, bn, mr, zh, my, ko, ja, mn, ta\\}}} \\
    \hline 
    & & \multicolumn{10}{c||}{\thead{Clustering Type 1\\}} & \multicolumn{10}{c||}{Clustering Type 2} & \multicolumn{11}{c||}{Clustering Type 3} & 
    \multicolumn{3}{c}{Clustering Type 4}\\
    \hline
    &  & \textcolor{c11}{$c_{1}$} &  \textcolor{c12}{$c_{2}$} &\textcolor{c13}{$c_{3}$} & \textcolor{c14}{$c_{4}$} &\textcolor{c15}{$c_{5}$} & \textcolor{c16}{$c_{6}$} & \textcolor{c17}{$c_{7}$}& \textcolor{c18}{$c_{8}$}&\textcolor{c19}{$c_{9}$}& \textcolor{c110}{$c_{10}$}&
    \textcolor{c21}{\thead{$c_{1}$\\}} &  \textcolor{c22}{$c_{2}$} &\textcolor{c23}{$c_{3}$} & \textcolor{c24}{$c_{4}$} &\textcolor{c25}{$c_{5}$} & \textcolor{c26}{$c_{6}$} & \textcolor{c27}{$c_{7}$}& \textcolor{c28}{$c_{8}$}&\textcolor{c29}{$c_{9}$}& \textcolor{c210}{$c_{10}$}&
    \textcolor{c31}{$c_{1}$} &  \textcolor{c32}{$c_{2}$} &\textcolor{c33}{$c_{3}$} & \textcolor{c34}{$c_{4}$} &\textcolor{c35}{$c_{5}$} & \textcolor{c36}{$c_{6}$} & \textcolor{c37}{$c_{7}$}& \textcolor{c38}{$c_{8}$}&\textcolor{c39}{$c_{9}$}& \textcolor{c310}{$c_{10}$}& \textcolor{c311}{$c_{11}$}&
    \textcolor{c41}{$c_{1}$} &  \textcolor{c42}{$c_{2}$} &\textcolor{c43}{$c_{3}$}\\
    \midrule
    \multirow{5}{*}{\rotatebox{90}{\small{\textbf{IE/Armenian}}}}
     & & & & & & & & & & & & & & & & & & & & & & & & & & & & & & & & & & & \\
      & & & & & & & & & & & & & & & & & & & & & & & & & & & & & & & & & & & \\
    & hy  & - & - & - & - & - & \textcolor{c16}{13.7}  & - &  - & - & - &
    - & - & \textcolor{c23}{ 12.7} & - & - & - & - & - & - & - & 
    - & \textcolor{c32}{ 10.8} & - & - & - & - & - & - & -  & - & - & 
    \textcolor{c41}{\textbf{\underline{17.1}}} & -  & - \\ 
     & & & & & & & & & & & & & & & & & & & & & & & & & & & & & & & & & & & \\
      & & & & & & & & & & & & & & & & & & & & & & & & & & & & & & & & & & & \\
     \midrule
    \multirow{5}{*}{\rotatebox{90}{\small{\textbf{Kartvelian}}}}
     & & & & & & & & & & & & & & & & & & & & & & & & & & & & & & & & & & & \\
      & & & & & & & & & & & & & & & & & & & & & & & & & & & & & & & & & & & \\
    & ka  & - & - & \textcolor{c13}{ 9.2} & - & - & -  & - & - & - & - &
    - & - & - & - & \textcolor{c25}{\textbf{\underline{10.8}}} & - & - & - & - & - & 
    - & \textcolor{c32}{9.1} & - & - & - & - & - & - & -  & - & - & 
    \textcolor{c41}{8.8} & -  & - \\ 
     & & & & & & & & & & & & & & & & & & & & & & & & & & & & & & & & & & & \\
      & & & & & & & & & & & & & & & & & & & & & & & & & & & & & & & & & & & \\
      
     \midrule
    \multirow{5}{*}{\rotatebox{90}{\small{\textbf{Mongolic}}}}
     & & & & & & & & & & & & & & & & & & & & & & & & & & & & & & & & & & & \\
      & & & & & & & & & & & & & & & & & & & & & & & & & & & & & & & & & & & \\
    & mn  & \textcolor{c18}{ 5.7}  & - & - & - & - & -  & - & - & - & - &
    - & - & - & \textcolor{c24}{\textbf{\underline{6.6}}} & - & - & - & - & - & - & 
    - & - & \textcolor{c33}{5.3} & - & - & - & - & - & -  & - & - & 
    - & -  & \textcolor{c43}{6.2} \\ 
     & & & & & & & & & & & & & & & & & & & & & & & & & & & & & & & & & & & \\
      & & & & & & & & & & & & & & & & & & & & & & & & & & & & & & & & & & & \\
      
     \midrule
    \multirow{5}{*}{\rotatebox{90}{\small{\textbf{Dravidian}}}}
     & & & & & & & & & & & & & & & & & & & & & & & & & & & & & & & & & & & \\
      & & & & & & & & & & & & & & & & & & & & & & & & & & & & & & & & & & & \\
    & ta  & - & - & \textcolor{c13}{4.0} & - & - & -  & - & - & - & - &
    - & - & - & - & \textcolor{c25}{\textbf{\underline{5.7}}} & - & - & - & - & - & 
    - & - & \textcolor{c33}{3.9} & - & - & - & - & - & -  & - & - & 
    - & -  & \textcolor{c43}{3.4}  \\ 
     & & & & & & & & & & & & & & & & & & & & & & & & & & & & & & & & & & & \\
      & & & & & & & & & & & & & & & & & & & & & & & & & & & & & & & & & & & \\
      
     \midrule
    \multirow{5}{*}{\rotatebox{90}{\small{\textbf{Isolate}}}}
     & & & & & & & & & & & & & & & & & & & & & & & & & & & & & & & & & & & \\
      & & & & & & & & & & & & & & & & & & & & & & & & & & & & & & & & & & & \\
    & eu  & - & - & - & \textcolor{c14}{ 9.0} & - & -  & - & - & - & - &
    - & \textcolor{c22}{ 9.7} & - & - & - & - & - & - & - & - & 
   - & - & \textcolor{c33}{8.1} & - & - & - & - & - & -  & - & - & 
    \textcolor{c41}{\textbf{\underline{11.0}}} & -  & - \\ 
     & & & & & & & & & & & & & & & & & & & & & & & & & & & & & & & & & & & \\
      & & & & & & & & & & & & & & & & & & & & & & & & & & & & & & & & & & & \\
    \bottomrule
\end{tabular} 
}}
\caption{The results of clustering-based multilingual NMT with knowledge distillation for languages belong to IE/Armenian, Kartvelian, Mongolic, Dravidian, and Isolate families.}
\label{tab:clustering-4}
\end{sidewaystable*}

\end{document}